\documentclass[twoside]{article}
\usepackage[accepted]{aistats2025}

\usepackage{amsfonts}       % blackboard math symbols
\usepackage{nicefrac}       % compact symbols for 1/2, etc.
\usepackage{microtype}      % microtypography
\usepackage{xcolor}         % colors
\usepackage{subcaption} % for subfigures
\usepackage{afterpage}
\usepackage{placeins}
\usepackage{multirow}
\usepackage{float} % Add this in the preamble
\usepackage{placeins}

\usepackage{comment}
\usepackage[accepted]{aistats2025}
\usepackage[round]{natbib}

\usepackage{bm}
\usepackage{amsmath}
\usepackage{amssymb}
\usepackage{mathtools}
\usepackage{amsthm}
\usepackage{xcolor}

\usepackage{algorithm}
\usepackage{algorithmic}

\usepackage{booktabs} % For professional looking tables
\usepackage{hyperref} % For hyperlinks

\definecolor{darkred}{rgb}{0.55, 0, 0}

\usepackage{tikz}
\usetikzlibrary{automata, positioning, arrows.meta}

\tikzset{
    state/.style={
        circle,
        draw,
        minimum size=1cm
    }
}

\DeclareMathOperator*{\argmax}{arg\,max}
\DeclareMathOperator*{\argmin}{arg\,min}

\def\Hc{\mathcal{H}}
\def\Sc{\mathcal{S}}
\def\Ac{\mathcal{A}}
\def\Zc{\mathcal{Z}}

\def\Oc{\mathcal{O}}

\def\Vc{\mathcal{V}}

\def\Ec{\mathcal{E}}
\def\Dc{\mathcal{D}}
\def\Vc{\mathcal{V}}

\def\Oct{\tilde{\mathcal{O}}}

\def\E{\mathbb{E}}

\def\Rr{\mathbb{R}}
\def\Nn{\mathbb{N}}

\def\Zz{\mathbb{Z}}

\def\psibar{\overline{\psi}}

\def\nn{\nonumber}

\newtheorem{definition}{Definition}
\newtheorem{assumption}{Assumption}
\newtheorem{lemma}{Lemma}
\newtheorem{theorem}{Theorem}
\newtheorem{remark}{Remark}
\newtheorem{corollary}{Corollary}

\definecolor{commentcolor}{RGB}{255, 0, 0}

\newcommand{\aya}[1]{#1}

\begin{document}



\twocolumn[

\aistatstitle{Near-Optimal Sample Complexity in Reward-Free Kernel-Based Reinforcement Learning}

\aistatsauthor{Aya Kayal$^{1}$ \And Sattar Vakili$^{2}$ \And  Laura Toni$^{1}$ \And Alberto Bernacchia$^{2}$ }

\aistatsaddress{ $^{1}$University College London \And $^{2}$MediaTek Research } ]

\begin{abstract}
  Reinforcement Learning (RL) problems are being considered under increasingly more complex structures. While tabular and linear models have been thoroughly explored, the analytical study of RL under nonlinear function approximation, especially kernel-based models, has recently gained traction for their strong representational capacity and theoretical tractability. In this context, we examine the question of statistical efficiency in kernel-based RL within the reward-free RL framework, specifically asking: \emph{how many samples are required to design a near-optimal policy?}
    Existing work addresses this question under restrictive assumptions about the class of kernel functions. We first explore this question by assuming a \emph{generative model}, then relax this assumption at the cost of increasing the sample complexity by a factor of $H$, the length of the episode.  We tackle this fundamental problem using a broad class of kernels and a simpler algorithm compared to prior work. Our approach derives new confidence intervals for kernel ridge regression, specific to our RL setting, which may be of broader applicability. We further validate our theoretical findings through simulations.
\end{abstract}

\section{INTRODUCTION}

Reinforcement Learning (RL) with nonlinear function approximation is a powerful method for learning general Markov Decision Processes (MDPs) through interactions with the environment. Kernel ridge regression for the prediction of the expected value function is perhaps one of the most versatile methods that has gained traction in recent years~\citep{yang2020provably, vakili2024kernelized, chowdhury2023value}, and lends itself to theoretical analysis. As a burgeoning research area, there are still numerous open problems and challenges in this topic.

We focus our work on statistical aspects of RL within the reward-free RL framework \citep{jin2020reward,wang2020reward,qiu2021reward}, which involves an exploration phase and a planning phase. In the exploration phase, the reward is unknown; the algorithm interacts with the environment to gather information about the underlying MDP, in the form of a dataset of transitions. In the planning phase, the reward is revealed; the algorithm uses the knowledge of the reward and the dataset gathered in the exploration phase to design a near-optimal policy. The planning phase is thus akin to offline RL~\citep{precup2000eligibility,antos2008learning,munos2008finite, levine2020offline,xie2021bellman,chen2019information}. In this paper, we answer the following fundamental question:  %The fundamental question we ask is: 
\emph{Under some reasonable assumptions on the underlying MDP, what is the minimum number of samples required to enable designing a near-optimal policy?}

We refer to the number of samples as \emph{sample complexity} and measure the optimality of the eventual policy in terms of error in the value function. In particular we refer to a policy as $\epsilon$-optimal if its value function is at most a small $\epsilon>0$ away from that of the optimal policy for all states. 

The reward-free RL framework has been studied in tabular \citep{jin2020reward} and linear~\citep{wang2020reward,hu2022towards,wagenmaker2022reward} settings.
% , with results shown in Table~\ref{samplecomplexitytable}.
Under the tabular setting, it has been shown that $\Oc(|\Sc|^2|\Ac|H^5/\epsilon^2)$ samples are sufficient to achieve an $\epsilon$-optimal policy, where $\Sc$ and $\Ac$ are the state and action spaces, respectively, and $H$ represents the length of episode. In the linear setting, a sample complexity of $\Oc(d^3H^6/\epsilon^2)$ has been established that does not scale with the size of the state-action space, but the ambient dimension $d$ of the linear model representing the transition structure of the MDP. With the limitations of the linear model~\citep[e.g., as shown in][]{lee2023demystifying}, recent works have considered non-linear function approximation in RL. The work
of~\citet{qiu2021reward} considered the reward-free RL framework with 
kernel-based function approximation. However, their results only apply to very smooth kernels with exponential eigendecay, such as Squared Exponential (SE), but fail to provide finite sample complexity applicable to a large class of kernels of interest with polynomial eigendecay (see Definition~\ref{def:eigendecay}), such as Mat{\'e}rn family or Neural Tangent (NT) kernels.
This shortcoming arises from the bias in the collected samples. Specifically in the exploration phase of~\citet{qiu2021reward}, the samples are adaptively collected to achieve a high value with respect to a hypothetical reward ---proportional to the uncertainties of the kernel ridge regression--- introducing bias to the samples and inflating confidence intervals. 
%\aya{Another closely related work addressing reward-free RL in the kernel setting is \citep{vakilireward}. Similar to \citep{qiu2021reward}, it uses a hypothetical reward proportional to the uncertainty of kernel ridge regression. However, it improves on \citep{qiu2021reward} by deriving order-optimal sample complexities for kernels with polynomially decaying eigenvalues, whereas the results of \citep{qiu2021reward} are unbounded for that case. This enhanced performance is achieved through an adaptive domain partitioning procedure inspired by \citep{vakili2024kernelized}. In this method, the state-action domain is adaptively divided into multiple subdomains as samples are collected, with kernel-based value function estimates constructed based on samples from the same subdomain, while discarding previous observations from other subdomains. More details about their method is given in Section~\ref{sec:comp}.
%They prove a regret bound of $\Oct\left((\frac{H^3}{\epsilon})^{2+\frac{2}{p-1}}\right)$.
%Although their approach offers theoretical advantages, it is tedious to implement in practice due to complex domain partitioning structure. Moreover, discarding samples may degrade the empirical performance, a concern that is not addressed in~\cite{vakilireward}. Additionally, their theoretical results depend on specific assumptions about the relationship between kernel eigenvalues and domain size, which limits generality of their work.}

Another closely related work on reward-free RL in the kernel setting is \citet{vakilireward}, which, like \citet{qiu2021reward}, uses a hypothetical reward proportional to the uncertainty of kernel ridge regression. However, it improves upon \citet{qiu2021reward} by providing order-optimal sample complexities for kernels with polynomially decaying eigenvalues, where \citet{qiu2021reward}'s results are unbounded. This is achieved via an adaptive domain partitioning procedure inspired by \citet{vakili2024kernelized}. In this method, the state-action domain is adaptively divided into multiple subdomains as samples are collected, with kernel-based value function estimates constructed based on samples from the same subdomain, while discarding previous observations from other subdomains. Although their approach offers theoretical advantages, it is tedious to implement in practice due to complex domain partitioning structure. Moreover, discarding samples may degrade the empirical performance, a concern that is not addressed in~\cite{vakilireward}. Additionally, their theoretical results depend on specific assumptions about the relationship between kernel eigenvalues and domain size, which limits generality of their work. A detailed comparison between our work and the two closely related works of~\citet{qiu2021reward} and~\citet{vakilireward} is provided in Appendix~\ref{appx:comp} along with a more comprehensive literature review in Appendix~\ref{appx:lit_review}.

In contrast to the existing work, this paper establishes near-optimal sample complexities for the reward-free kernel-based RL framework over a general class of kernels, without relying on restrictive assumptions. This is accomplished via a simple algorithm and a novel confidence interval for unbiased samples, broadly applicable to other RL settings (offline RL, model-based, infinite horizon), and supported by empirical evidence.
Specifically, we start with a case where a \emph{generative model}~\citep{kakade2003sample} is present and it permits the algorithm to sample state-actions of its choice during the exploration phase, not limiting the algorithm to stay on the Markovian trajectory. This setting has been extensively considered in previous work on statistical efficiency of RL~\citep[see, e.g.,][]{kearns1998,gheshlaghi2013minimax, sidford2018near, sidford2018variance, agarwal2020model,yang2019sample}. In the presence of a generative model, we propose a simple algorithm that collects \emph{unbiased} samples by choosing the state-actions with highest kernel-based regression uncertainty at each step. 
We derive order-optimal sample complexities for this algorithm in terms of $\frac{1}{\epsilon}$, while \citet{vakilireward} do not offer any particular advantages in the generative model case. 
Generative models are applicable in scenarios like games where the algorithm can manipulate the current state, offering insights into the statistical aspects of RL.  However, this may not be the case in other scenarios. Inspired by the analysis of the exploration algorithm with a generative model, we propose a second online exploration algorithm that collects samples adhering to the Markovian trajectory. We prove that this relaxing of generative model requirement incurs merely an $H$ factor increase in the sample complexity.%; implying that the price of online samples compared to generative samples is an $H$ factor. 

To highlight the significance of our results, we consider kernels with polynomial eigendecay %such as the Mat{\'e}rn family and NT kernels
that are of practical and theoretical interest~\citep{srinivas2009gaussian, jacot2018neural, vakili2023information}. When the eigenvalues of the kernel decay polynomially as $\Oc(m^{-p})$ ---see Definition~\ref{def:eigendecay}---the results of \citet{qiu2021reward} lead to possibly vacuous (infinite) sample complexities, while we prove an $\Oct((\frac{H^3}{\epsilon})^{2+\frac{2}{p-1}})$ sample complexity for the generative setting and $\Oct(H(\frac{H^3}{\epsilon})^{2+\frac{2}{p-1}})$ for the online setting. Our sample complexity results are comparable to those of \citet{vakilireward}. In a technical comparison, their approach requires a specific assumption on the dependence between kernel eigenvalues and domain size~\citep[see,][Definition $4.1$]{vakilireward}, which we do not. Additionally, they employ a sophisticated domain partitioning algorithm that is more difficult to implement and possibly inefficient in practice, whereas our algorithm is simpler and more straightforward.
In the case of Mat{\'e}rn kernel with smoothness parameter $\nu$ on a $d$-dimensional domain, where $p=1+\frac{2\nu}{d}$, our results translate to a sample complexity of $\Oct(H(\frac{H^3}{\epsilon})^{2+\frac{d}{\nu}})$, that matches the $\Omega((\frac{1}{\epsilon})^{2+\frac{d}{\nu}})$ lower bound proven in~\cite{scarlett2017lower} for the degenerate case of bandits with $H=1$. Our sample complexities thus are not generally improvable in their scaling with $\frac{1}{\epsilon}$. 

To achieve these results, we establish a confidence interval applicable to kernel ridge regression in our RL setting that may be of broader interest. The key technical novelties of this confidence interval involves leveraging the structure of RKHS and the properties of unbiased, independent samples. The main results regarding the confidence interval and sample complexities of the two exploration algorithms, with and without the generative model, are presented in Theorems~\ref{the:conf}, \ref{the:gen} and~\ref{the:main}, respectively, in Section~\ref{sec:anal}. We empirically validate our analytical findings through numerical experiments comparing the performance of our proposed exploration algorithms with that of \citet{qiu2021reward}, as detailed in Section~\ref{sec:exp}. Section~\ref{sec:pf} provides an overview of episodic MDPs, the reward-free RL framework, and kernel-based models. Section~\ref{sec:alg} presents our algorithms for both the exploration and planning phases. Detailed proofs of theorems, along with the details of experiments and further experimental results, are included in the appendix due to space limitations.
%While we here presented the most closely related work,   
%given the extensive body of research in RL, and due to space constraints, a broader overview of the related work is provided in Appendix \ref{sec:relatedworks}.
%In the introduction, we presented the most closely related work. A more comprehensive literature review is provided in Appendix~\ref{sec:relatedworks}.
\section{PRELIMINARIES AND PROBLEM FORMULATION}\label{sec:pf}

In this section, we introduce the episodic MDP setting, describe the reward-free RL framework, provide background on kernel methods, and outline our technical assumptions.

\subsection{Episodic MDP}
An episodic MDP can be described by the tuple $M=(\Sc,\Ac, H, P, r)$, where $\Sc$ denotes the state space, $\Ac$ the action space, and the integer $H$ the length of each episode. Here, $r=\{r_h\}_{h=1}^H$ represents the reward functions, and $P=\{P_h\}_{h=1}^H$ the transition probability distributions.\footnote{We deliberately do not use the standard term transition kernel for $P_h$, to avoid confusion with kernel in kernel-based learning.}
The state-action space is denoted by $\Zc=\Sc\times\Ac$. The notation $z=(s,a)$ is used throughout the paper for state-action pairs. For each step $h\in[H]$, the reward function $r_h: \Zc\rightarrow [0,1]$ is supposed to be deterministic for simplicity, and $P_h(\cdot|s,a)$ is the unknown transition probability distribution on $\Sc$ for the next state given the current state-action pair~$(s,a)$.

%\begin{figure}[ht]
%    \centering
%    \begin{tikzpicture}
%
%    \tikzset{state/.style={circle, draw=blue, minimum size=1.1cm, align=center}}
%        % Nodes
%        \node[line width=0.3mm,  blue, state] (s1) {\tcd{$s_1$}};
%        \node[line width=0.3mm,  blue, state, right=of s1] (s2) {\tcd{$s_2$}};
%        \node[right=of s2] (dots) {$\cdots$};
%        \node[line width=0.3mm,  blue, state, right=of dots] (sH) {\tcd{$s_H$}};
%        \node[dashed, line width=0.3mm,  blue, state, right=of sH] (sH1) {\tcd{$s_{H+1}$}};
%
%        % Edges
%        \draw[line width=0.3mm,  blue, ->] (s1) to[bend left] node[above] {\tcd{$a_1$}} (s2);
%        \draw[line width=0.3mm,  blue,->] (s2) to[bend left] (dots);
%        \draw[line width=0.3mm,  blue,->] (dots) to[bend left] node[above] {\tcd{$a_{H-1}$}} (sH);
%        \draw[dashed, line width=0.3mm,  blue,->] (sH) to[bend left] node[above] {\tcd{$a_{H}$}} (sH1);
%    \end{tikzpicture}
%    \caption{Illustration of an Episodic MDP with an episode of length~$H$.}
%    \label{fig:episodic_mdp}
%\end{figure}

A policy $\pi = \{\pi_h: \Sc \rightarrow \Ac\}_{h=1}^H$ determines the action $\pi_h(s)$ ---possibly random--- taken by the agent at state $s$ during each step $h$.  
At the beginning of each episode, the environment picks an arbitrary initial state $s_1$. The agent adopts a policy $\pi=\{\pi_h\}_{h=1}^H$. For each step $h\in[H]$, the agent observes the current state $s_h\in\Sc$, and selects an action $a_h=\pi_h(s_h)$. The subsequent state, $s_{h+1}$, is then drawn from the transition probability distribution $P_h(\cdot|s_h, a_h)$. The episode ends when the agent receives the final reward $r_H(s_H,a_H)$.

We are interested in maximizing the expected total reward in the episode, starting at step $h$. This is quantified by the value function, which is defined as follows:
\begin{equation}
V^{\pi}_h(s) = \E\left[\sum_{h'=h}^Hr_{h'}(s_{h'},a_{h'})\bigg|s_{h}=s\right], \forall s\in\Sc, h\in[H],
\end{equation}
where the expectation is taken with respect to the randomness in the trajectory $\{(s_h,a_h)\}_{h=1}^H$ obtained by the policy~$\pi$. It can be shown that under mild assumptions (e.g., continuity of $P_h$, compactness of $\Zc$, and boundedness of $r$), there exists an optimal policy $\pi^{\star}$ which attains the maximum possible value of $V^{\pi}_{h}(s)$ at every step and at every state~\citep[e.g., see,][]{puterman2014markov}. We use the notation
$
V_{h}^{\star}(s) = \max_{\pi}V_h^{\pi}(s), ~\forall s\in\Sc, h\in[H]
$.
By definition $V^{\pi^{\star}}_h=V_{h}^{\star}$.
An $\epsilon$-optimal policy is defined as follows. 

\begin{definition}\label{def:epsopt}
($\epsilon$-optimal policy) For $\epsilon>0$, a policy $\pi$ is called $\epsilon$-optimal if it achieves near-optimal values from any initial state as follows:
$
V_1^{\pi}(s) \geq V_1^{\star}(s) - \epsilon, ~~~ \forall s \in \Sc.
$
\end{definition}

Policy design often relies on the expected value of a value function with respect to the transition probability distribution, presented using the following notation:
\begin{equation}
    [P_hV](s,a) := \E_{s'\sim P_h(\cdot|s,a)}[V(s')].  
\end{equation}

We also define the state-action value function $Q^{\pi}_h:\Zc\rightarrow [0,H]$ as follows:
\begin{equation}
Q_h^{\pi}(s,a) = \E_{\pi}\left[
\sum_{h'=h}^Hr_{h'}(s_{h'},a_{h'})\bigg|s_h=s, a_h=a
\right],
\end{equation}
where the expectation is taken with respect to the randomness in the trajectory $\{(s_h,a_h)\}_{h=1}^H$ obtained by the policy~$\pi$.
The Bellman equation associated with a policy $\pi$ is then represented as
\begin{align*}
Q_h^{\pi}(s,a) = r_h(s,a) + [P_hV^{\pi}_{h+1}](s,a), \\
V_h^{\pi}(s) = \E[Q_h^{\pi}(s,\pi_h(s))],~~
V_{H+1}^{\pi} = \bm{0}.
\end{align*}
The notation $V=\bm{0}$ is used for $V(s)=0$, for all $s\in\Sc$.
We may specify the reward function in $V^{\pi}, Q^{\pi}, V^{\star}, Q^{\star}$ notations for clarity, for example, $V^{\pi}(s;r)$ and $Q^{\star}(z;r)$.

\subsection{Reward-Free RL Framework}

We aim to learn $\epsilon$-optimal policies while minimizing the samples collected during exploration.
%the smallest possible number of interactions with the environment during exploration episodes. 
Specifically, we employ the reward-free RL framework, which consists of two phases: {exploration} and {planning}. In the exploration phase, we collect
a dataset $\Dc_{N}=\{\Dc_{h,N}\}_{h\in[H]}$, where each $\Dc_{h,N}=\left\{\left(s_{h,n}, a_{h,n}, s'_{h+1,n}\sim P_h(\cdot|s_{h,n},a_{h,n})\right)\right\}_{n\in[N]}$ consists of $N$ transition samples at step~$h$. Then, in the planning phase, once the reward $r$ is revealed, we design a policy specific to reward $r$ using the data collected during the exploration phase. The number $N$ denotes the sample complexity required to design an $\epsilon$-optimally performing policy. A critical question arises: \emph{How many exploration episodes are necessary to achieve $\epsilon$-optimal policies?} We provide an answer in this work. 

\subsection{Kernel Ridge Regression}

A main step in RL with function approximation, keeping the Bellman equation in mind, is to derive statistical predictions and bounds for the expected value function
$[PV]:\Zc\rightarrow \Rr$, for some given value function $V:\Sc\rightarrow \Rr$ and conditional probability distribution $P(\cdot|z)$. Let us use the notation $f=[PV]$. Suppose that we are given $n$ noisy observations of $f$, represented as $\left\{(z_i, y_i)\right\}_{i\in[n]}$, where $y_i=f(z_i)+\varepsilon_i$, and $\varepsilon_i$ denotes zero-mean random noise. Provided a positive definite kernel $k: \Zc\times\Zc\rightarrow \Rr$ and employing kernel ridge regression, we can make the following prediction for $f$:
\begin{equation}
    \hat{f}_n(z) = k^{\top}_n(z)(K_n+\tau^2 I)^{-1}\bm{y}_n,
\end{equation}
where $k_n(z)=[k(z,z_1), k(z,z_2), \cdots, k(z,z_n)]^{\top}$ is the pairwise kernel values between $z$ and observation points, $K_n=[k(z_i, z_j)]_{i,j\in[n]}$ is the Gram matrix, $\bm{y}_n=[y_1, y_2, \cdots, y_n]^{\top}$ is the vector of observations, $\tau>0$ is a free parameter, and $I$ is the identity matrix, appropriately sized to match the dimensions of $K_n$. In addition, the following uncertainty estimate can be utilized to bound the prediction error:
\begin{equation}\label{eq:krrvar}
    \sigma_n^2(z)= k(z,z)-k^{\top}_n(z)(K_n+\tau^2 I)^{-1}k_n(z)
\end{equation}
In particular, various $1-\delta$ confidence intervals of the form $|f(z)-\hat{f}_n(z)|\le \beta(\delta) \sigma_n(z)$, under various assumptions, are proven, where $\beta(\delta)$ is a confidence interval width multiplier that depends on the setting and assumptions~\citep{abbasi2013online, chowdhury2017kernelized, vakili2021optimal,  whitehouse2024sublinear}.
One of our primary contributions is establishing a novel confidence intervals for $f=[PV]$, applicable to our RL setting.
Equipped with the confidence intervals, we are able to design policies using least squares value iteration or its \emph{optimistic} variant.

\paragraph{Reproducing Kernel Hilbert Spaces and Mercer Representation:} Mercer theorem provides a representation of a positive definite kernel function $k$ using an infinite-dimensional feature map: $k(z,z')=\sum_{m=1}^{\infty}\gamma_m\varphi_m(z)\varphi_m(z')$,
% \begin{eqnarray}\label{eq:Mercer}
% k(z,z')=\sum_{m=1}^{\infty}\gamma_m\varphi_m(z)\varphi_m(z'),
% \end{eqnarray}
where $\gamma_m > 0$ are referred to as Mercer eigenvalues, and $\varphi_m$ are the corresponding eigenfunctions. The Reproducing Kernel Hilbert Space (RKHS) associated with $k$, denoted $\Hc_k$, is defined as:
$
    \Hc_k =\{f: f =\sum_{m=1}^{\infty}w_m\phi_m,~~~ w_m\in\Rr,~ \|\bm{w}\|<\infty\},
$
where $\phi_m := \sqrt{\gamma_m} \varphi_m$ form an orthonormal basis of $\Hc_k$. Here, $\bm{w} = [w_1, w_2, \cdots]^{\top}$ represents a possibly infinite-dimensional weight vector. The RKHS norm of $f$ is defined as $\|f\|_{\Hc_k}=\|\bm{w}\|$, the $\ell^2$ norm of the weight vector. Formal statements and further details are provided in Appendix~\ref{appx:rkhs}.

To effectively use the confidence intervals established by the kernel-based models on $f$,
we require the following assumption.
\begin{assumption}\label{ass:rkhsnorm}
    We assume $P_h(s|\cdot, \cdot)\in \Hc_k$, for some positive definite kernel $k$, and $\|P_h(s|\cdot, \cdot)\|_{\Hc_k}\le 1$, for all $s\in \Sc$ and $h\in[H]$. 
\end{assumption}

Consequently, for all $V:\Sc\rightarrow [0,H]$, we have $\|[PV]\|_{\Hc_k}=\Oc(H)$. See~\cite{yeh2023sample}, Lemma~$3$,
for a proof.

\paragraph{Information Gain and Eigendecay:}
The analytical results in kernel-based RL and bandits are often given in terms of a kernel specific complexity term referred to as maximum information gain, defined as follows~\citep{srinivas2009gaussian, vakili2021information}:
\begin{equation}
    \Gamma(n)=\sup_{\{z_i\}_{i=1}^{n}\subset \Zc}\frac{1}{2}\log\det(I+\frac{K_n}{\tau^2}),
\end{equation}
Maximum information gain depends on the eigendecay defined as follows. 
\begin{definition}\label{def:eigendecay}
    A kernel $k$ is said to have a polynomial (resp. exponential) eigendecay if $\gamma_m=\Oc(m^{-p})$ (resp. $\gamma_m=\Oc(c^{m})$), for some $p>1$ ($c<1$), where $\gamma_m$ are the Mercer eigenvalues in decreasing order. 
\end{definition}
For kernels with polynomial and exponential eigendecays, $\Gamma(n) = \Oct(n^{\frac{1}{p}})$ and $\Gamma(n) = \Oc(\text{polylog}(n))$, respectively~\citep{vakili2021information}.

\section{ALGORITHM DESCRIPTION}
\label{sec:alg}

We now present our algorithms for both the exploration and planning phases. We begin by presenting the algorithm for the planning phase, as it remains unchanged across various exploration algorithms.
\begin{algorithm}[ht]
   \caption{Planning Phase}
   \label{alg:plan}
\begin{algorithmic}
   \STATE {\bfseries Input:} $\tau$, $\beta$, $\delta$, $k$, $M(\Sc,\Ac, H, P, r )$, and exploration dataset $\Dc_{N}$.
   \FOR{$h=H, H-1, \cdots, 1,$}
        \STATE Compute the prediction $\hat{g}_h$ according to~\eqref{eq:ghn};
        \STATE Let $Q_h(\cdot, \cdot) = \Pi_{[0,H]}[\hat{g}_h(\cdot, \cdot)+r_h(\cdot, \cdot) + \beta(\delta) \sigma_{h,N}(.,.) ] $;
        \STATE $V_h(\cdot)= \max_{a\in\Ac}Q_h(\cdot, a)$;
        \STATE $\pi_h(\cdot) = \argmax_{a\in\Ac}Q_h(\cdot,a)$;
   \ENDFOR
   \STATE {\bfseries Output:} $\{\pi_h\}_{h\in[H]}$. 
\end{algorithmic}
\end{algorithm}

\begin{algorithm}[ht]
\caption{Exploration Phase \textbf{with} Generative Model}\label{alg:exp_gen}
\begin{algorithmic}[1]
\REQUIRE $\tau$, $k$, $\Sc$, $\Ac$, $H$, $P$, $N$;
\STATE Initialize $\Dc_{h,0}=\{\}$, for all $h\in[H]$;
\FOR{$n=1,2,\cdots, N$}
    \FOR{$h=1,2,\cdots, H$}
    \STATE Let $s_{h,n}, a_{h,n} = \argmax_{s \in\Sc, a\in\Ac}\sigma_{h, n-1}(s,a)$;
    \STATE Observe $s'_{h+1,n}\sim P_h(\cdot|s_{h,n}, a_{h,n})$;
    \STATE Update $\Dc_{h,n} = \Dc_{h,n-1} \bigcup\{s_{h,n}, a_{h,n}, s'_{h+1,n}\}$.
    \ENDFOR
\ENDFOR
\STATE {\bfseries Output:} $\Dc_{N}$. 
\end{algorithmic}
\end{algorithm}

\begin{algorithm}[ht]
\caption{Exploration Phase \textbf{without} Generative Model}\label{alg:exp2}
\begin{algorithmic}
\REQUIRE $\tau$, $k$, $\beta$, $\delta$, $\Sc$, $\Ac$, $H$, $P$, $N$;
\STATE Initialize $\Dc_{h,0}=\{\}$, for all $h\in[H]$;
\FOR{$n=1,2,\cdots, N$}
    \FOR{$h_0=1, 2,\cdots, H$}
    \STATE Initialize $V_{h_0+1,n}=\bm{0}$
    \FOR{$h=h_0, h_0-1, \cdots, 1$}
        \STATE Obtain $\hat{f}_{h,(n,h_0)}$; $Q_{h, (n,h_0)}$, and $V_{h,(n,h_0)}(\cdot)$ according to~\eqref{eq:mean_predictor} and ~\eqref{eq:value_func}, respectively. 
    \ENDFOR
    \FOR{$h=1,2,\cdots, h_0$}
    \STATE Observe $s_{h,n}$; Take action $a_{h,n} = \argmax_{a\in\Ac}Q_{h, n}(s_{h,n},a)$; 
    \ENDFOR
    \STATE Update $\Dc_{h_0, n} = \Dc_{h_0,n-1} \bigcup\{s_{{h_0},n}, a_{h_0,n}, s_{h_0+1,n}\}$
    \ENDFOR
\ENDFOR
\end{algorithmic}
\end{algorithm}

\subsection{Planning Phase} In the planning phase, the reward function $r$ is revealed to the learner. In addition, a dataset $\Dc_N=\{\Dc_{h,N}\}_{h\in[H]}$ is available, with $\Dc_{h,N}=\{s_{h,n}, a_{h,n}, s'_{h+1,n}\sim P_h(\cdot|s_{h,n}, a_{h,n})\}_{ n\in[N]}$ for each step $h\in[H]$. The objective is to leverage the knowledge of the reward function and utilize the dataset to design a near-optimal policy.
As mentioned in the introduction, the planning phase comprises of an offline RL design without further interaction with the environment.

In the planning phase of our algorithm, we derive a policy using least squares value iteration. Specifically, at step $h$, we compute a prediction, $\hat{g}_{h}$, for the expected value function in the next step $[P_hV_{h+1}]$. We then define
\aya{\begin{equation}\label{eq:Qplan}
    Q_h(\cdot, \cdot) = \Pi_{[0,H]}\left[r_h(\cdot, \cdot)+ \hat{g}_h(\cdot, \cdot)  +  \beta(\delta) \sigma_{h,N}(\cdot,\cdot)\right],
\end{equation}
}
where $\Pi_{[a,b]}$ denotes projection on $[a,b]$ interval. 
The policy $\pi$ is then obtained as a greedy policy with respect to $Q$. For each $h\in[H]$,
\begin{equation*}
    \pi_h(\cdot) = \argmax_{a\in\Ac}Q_h(\cdot, a).
\end{equation*}
We now detail the computation of $\hat{g}_h$. Keeping the Bellman equation in mind and starting with $V_{H+1}=\bm{0}$, $\hat{g}_h$ is the kernel ridge predictor for $[P_hV_{h+1}]$. This prediction uses $N$ observations $\bm{y}_{h}=[V_{h+1}(s'_{h+1, 1}),V_{h+1}(s'_{h+1, 2}), \cdots,V_{h+1}(s'_{h+1, N}) ]^{\top}$ at points $\{z_{h,n}\}_{n=1}^N$. 
Recall that 
$
\E_{s'\sim P(\cdot|z_{h, n})}\left[V
_{h+1}(s')  \right]  = [P_hV_{h+1}](z_{h,n})
$.
The observation noise $V_{h+1}(s'_{h+1,n})-[P_hV_{h+1}](z_{h,n})$ is due to random transitions and is bounded by $H-h\le H$. Specifically, 
\begin{equation}\label{eq:ghn}
    \hat{g}_h(z) = k^{\top}_{h,N}(z)(\tau^2 I + K_{h,N})^{-1} \bm{y}_{h},
\end{equation}
where $k_{h,N}(z)=[k(z,z_{h,1}), k(z,z_{h,2}), \cdots, k(z,z_{h,N})]^{\top}$ is the pairwise kernel values between $z$ and observation points and $K_{h,N}=[k(z_{h,i}, z_{h,j})]_{i,j\in[N]}$ is the Gram matrix. 
\aya{Also, $\sigma_{h,N}$ in~\eqref{eq:Qplan} is specified as follows:
\begin{equation}
   \sigma^2_{h,N}(z) = k(z,z) -k^{\top}_{h,N}(z)(\tau^2I+K_{h,N})^{-1}k_{h,N}(z). 
\end{equation}
}
We then define $Q_h$ according to~\eqref{eq:Qplan} and also set
\begin{equation*}
    V_h(s)=\max_{a\in\Ac} Q_h(s, a).
\end{equation*}
The values of $\hat{g}_h$, \aya{$ \sigma_{h,N}$, }$Q_h$ and $V_h$ are obtained recursively for $h=H,H-1, \cdots, 1$.
For a pseudocode, see Algorithm~\ref{alg:plan}.

% \sattar{Comment on how this is slightly different from Qiu et al. }

%\begin{algorithm}[tb]
%   \caption{Planning Phase}
%   \label{alg:plan}
%\begin{algorithmic}
%   \STATE {\bfseries Input:} $\tau$, $\beta$, $\delta$, $k$, $M(\Sc,\Ac, H, P, r )$, and exploration data set $\Dc_{N}$.
%   \FOR{$h=H, H-1, \cdots, 1,$}
%        \STATE Compute the prediction $\hat{g}_h$ according to~\eqref{eq:ghn};
%        \STATE Let $Q_h(\cdot, \cdot) = \Pi_{[0,H]}[\hat{g}_h(\cdot, \cdot)+r_h(\cdot, \cdot)] $;
%        \STATE $V_h(\cdot)= \max_{a\in\Ac}Q_h(\cdot, a)$;
%        \STATE $\pi_h(\cdot) = \argmax_{a\in\Ac}Q_h(\cdot,a)$;
%   \ENDFOR
%   \STATE {\bfseries Output:} $\{\pi_h\}_{h\in[H]}$. 
%\end{algorithmic}
%\end{algorithm}

\subsection{Exploration Phase} In the exploration phase, the algorithm collects a dataset
$\Dc_{N}=\{\Dc_{h,N}\}_{h\in[H]}$, where
$\Dc_{h,N}=\{s_{h,n},a_{h,n}, s'_{h+1,n}\}_{h\in[H], n\in[N]}$ for each $h\in[H]$, %. As we showed, these observations will be 
later used in the planning phase to design a near-optimal policy. The primary goal during this phase is to gather the most informative observations.

% As discussed in the introduction, existing work fails to achieve order-optimal or even finite sample complexities without imposing strong and restrictive assumptions %such as assuming a kernel with exponential eigendecay.
Initially, we consider a preliminary case where a \emph{generative model}~\citep{kakade2003sample} is present that can produce transitions for the state-actions selected by the algorithm. Under this setting, we demonstrate that a simple rule for data collection leads to improved and desirable sample complexities. Inspired by these results, we introduce a novel algorithm that completely relaxes the requirement for a generative model, at the price of increasing the number of exploration episodes by a factor of $H$. The key aspect of our algorithms is the \emph{unbiasedness}--statistical independence of the collected samples, which means that the observation points do not depend on previous transitions.

\subsubsection{Exploration with a Generative Model}

In this section, we outline the exploration phase when a generative model is present. At each step $h$ of the current exploration episode, uncertainties derived from kernel ridge regression are employed to guide exploration. Specifically, let
\begin{equation} \label{eq:std_dev}
   \sigma^2_{h,n}(z) = k(z,z) -k^{\top}_{h,n}(z)(\tau^2I+K_{h,n})^{-1}k_{h,n}(z) 
\end{equation}
where $k_{h,n}(z) = [k(z,z_{h,1}),k(z,z_{h,2}), \cdots, k(z,z_{h,n}) ]^{\top}$ is the vector of kernel values between the state-action of interest and past observations in $\Dc_{h,n}$, and $K_{h,n}=[k(z_{h,i}, z_{h,j})]_{i,j=1}^n$ is the Gram matrix of pairwise kernel values between past observations in $\Dc_{h,n}$. Equipped with $\sigma_{h,n}(z)$, at step $h$, we select
\begin{equation}\label{eq:sel_rule}
    s_{h,n}, a_{h,n}=\argmax_{s\in\Sc,a\in\Ac} \sigma_{h,n-1}(s,a),
\end{equation}
and observe the next state $s'_{h+1,n}\sim P_h(\cdot|s_{h,n}, a_{h,n})$. We then add this data point to the dataset and update $\Dc_{h,n}=\Dc_{h,n-1}\cup\{(s_{h,n}, a_{h,n}, s'_{h+1,n})\}$. For a pseudocode, see Algorithm~\ref{alg:exp_gen}.

%\begin{algorithm}[ht]
%\caption{Exploration Phase \textbf{with} Generative Model}\label{alg:exp_gen}
%\begin{algorithmic}[1]
%\REQUIRE $\tau$, $k$, $\Sc$, $\Ac$, $H$, $P$, $N$;
%\STATE Initialize $\Dc_{h,0}=\{\}$, for all $h\in[H]$;
%\FOR{$n=1,2,\cdots, N$}
%    \FOR{$h=1,2,\cdots, H$}
%    \STATE Let $s_{h,n}, a_{h,n} = \argmax_{s,a\in\Ac}\sigma_{h, n-1}(s,a)$;
%    \STATE Observe $s'_{h+1,n}\sim P_h(\cdot|s_{h,n}, a_{h,n})$;
%    \STATE Update $\Dc_h^n \bigcup\{s_{h,n}, a_{h,n}, s'_{h+1,n}\}$.
%    \ENDFOR
%\ENDFOR
%\STATE {\bfseries Output:} $\Dc_{N}$. 
%\end{algorithmic}
%\end{algorithm}
We highlight that the selection rule~\eqref{eq:sel_rule} relies on the generative model that allows the algorithm to deviate from the Markovian trajectory and move to a state of its choice. Since observations $(s_{h,n}, a_{h,n})$ are selected based on maximizing $\sigma_{h,n-1}$, which by definition \eqref{eq:std_dev} does not depend on previous transitions $\{s'_{h+1,i}\}_{i=1}^{n-1}$,  the statistical independence conveniently holds. The generative model setting is feasible in contexts such as games, where the player can manually set the current state. However, this may not always be possible in other scenarios. Next, we introduce our online algorithm, which strictly stays on the Markovian trajectory.

\subsubsection{Exploration without Generative Models}

In this section, we show that a straightforward algorithm, in contrast to existing approaches, achieves near-optimal performance in an online setting without requiring a generative model. Compared to the scenario with a generative model, the sample complexity of this algorithm increases by a factor of $H$. For a detailed and technical comparison with existing work, please refer to Appendix~\ref{appx:comp}. %, highlighting how it may lead to trivial (infinite) sample complexities.

Our online algorithm operates as follows: in each exploration episode, only one data point specific to a step $h$ is collected ---this accounts for the $H$ scaling in sample complexity. This observation however is collected in an unbiased way, which eventually leads to tighter performance guarantees. Specifically, at episode $nH+h_0$, where $n\in[N]$ and $h_0\in[H]$, the algorithm collects an informative sample for the transition at step $h_0$. This results in a total of $N$ samples at each step over $NH$ episodes. The algorithm initializes $V_{h_0+1,(n,h_0)}=\bm{0}$.
Let $\hat{f}_{h,(n,h_0)}$ and $\sigma_{h,n}$ represent the predictor and uncertainty estimator for $[P_hV_{h+1,(n,h_0)}]$, respectively. These are derived from the historical data $\Dc_{h,n-1}$ of observations at step $h$. Specifically, 
\begin{align}\nn
    \hat{f}_{h,(n,h_0)}(z) &= k^{\top}_{h,n}(z)(K_{h,n}+\tau^2 I)^{-1}\bm{y}_{h,n},  \\
    \sigma^2_{h,n}(z) &= k(z,z) - k^{\top}_{h,n}(z)(K_{h,n}+\tau^2 I)^{-1}k_{h,n}(z), \label{eq:mean_predictor}
\end{align}
where $k_{h,n}(z) = [k(z,z_{h,1}),k(z,z_{h,2}), \cdots, k(z,z_{h,n}) ]^{\top}$ is the vector of kernel values between the state-action of interest and past observations in $\Dc_{h,n}$, $K_{h,n}=[k(z_{h,i}, z_{h,j})]_{i,j=1}^n$ is the Gram matrix of pairwise kernel values between past observations in $\Dc_{h,n}$, and 
\begin{align*}
\bm{y}_{h,(n,h_0)} &= [V_{h+1,(n,h_0)}(s_{h+1,1}), V_{h+1,(n,h_0)}(s_{h+1,2}), \\
&\quad \cdots, V_{h+1,(n,h_0)}(s_{h+1,n}) ]^{\top}
\end{align*} 
is the vector of observations. 
We then have
%\begin{align}\nn 
%Q_{h,(n,h_0)}=\Pi_{0,H}\left[\hat{f}_{h,(n,h_0)}+\beta(\delta)\sigma_{h,n}\right], 
%~~~~~ \text{and} ~~~ V_{h,(n,h_0)}(\cdot) = \max_{a\in\Ac}Q_{h,(n,h_0)}(\cdot, a). 
%\end{align}
\begin{align*}
Q_{h,(n,h_0)} = \Pi_{0,H} \left[ \hat{f}_{h,(n,h_0)} + \beta(\delta) \sigma_{h,n} \right], \\
V_{h,(n,h_0)}(\cdot) = \max_{a \in \Ac} Q_{h,(n,h_0)}(\cdot, a) \label{eq:value_func}.
\end{align*}

%\begin{algorithm}[ht]
%\caption{Exploration Phase \textbf{without} Generative Model}\label{alg:exp2}
%\begin{algorithmic}
%\REQUIRE $\tau$, $k$, $\beta$, $\delta$, $\Sc$, $\Ac$, $H$, $P$, $N$;
%
%\FOR{$n=1,2,\cdots, N$}
%    \FOR{$h_0=1, 2,\cdots, H$}
%    \STATE Initialize $V_{h_0+1,n}=\bm{0}$
%    \FOR{$h=h_0, h_0-1, \cdots, 1$}
%        \STATE Obtain $\hat{f}_{h,n}$; $Q_{h, n}$, and $V_{h,n}(\cdot)$ according to~\eqref{eq:mean_predictor} and %~\eqref{eq:value_func}, respectively. 
%    \ENDFOR
%    \FOR{$h=1,2,\cdots, h_0$}
%    \STATE Observe $s_{h,n}$; Take action $a_{h,n} = \argmax_{a\in\Ac}Q_{h, n}(s_{h,n},a)$; 
%    \ENDFOR
%    \STATE Update $\Dc_{h_0}^n \bigcup\{s_{{h_0},n}, a_{h_0,n}, s_{h_0+1,n}\}$
%    \ENDFOR
%\ENDFOR
%\end{algorithmic}
%\end{algorithm}

The values of $Q_{h,(n,h_0)}$ and $V_{h,(n,h_0)}$ are obtained recursively for all $h\in[h_0]$. 
The exploration policy at episode $nH+h_0$ is then the greedy policy with respect to $Q_{h,(n-1,h_0)}$. The dataset is updated by adding the new observation to the dataset for step $h_0$, such that $\Dc_{h_0,n} = \Dc_{h_0,n-1}\cup \{(s_{h_0,n}, a_{h_0,n}, s_{h_0+1,n})\}$, while datasets for all other steps remain unchanged: $\Dc_{h,n} = \Dc_{h,n-1}$ for all $h\neq h_0$. This specific update ensures that the collected samples are unbiased. More specifically, the sample collected at $h_0$ solely relies on the uncertainty $\sigma_{h_0,n}$, due to the initialization $V_{h_0+1,(n,h_0)}=\bm{0}$ which implies $\hat{f}_{h_0,(n,h_0)}=\bm{0}$. Since $\sigma_{h_0,n}$ does not depend on previous transitions $s_{(h_0+1,i)}$ for any $i\leq n$, the samples at $h=h_0$ are unbiased. However, for \( h < h_0 \), the samples depend on both the uncertainty \( \sigma_{h,n} \) and the prediction \( \hat{f}_{h,(n,h_0)} \)~\eqref{eq:mean_predictor}. Since the prediction depends on the transitions $s_{(h+1,i)}$ for $i\leq n$, these samples are biased. As a result, we discard them and only retain the unbiased samples at \( h = h_0 \). This approach improves the rates in our analysis, albeit at the cost of a factor of~\( H \).

% A detailed technical comparison with the closely related works of~\cite{qiu2021reward} and~\cite{vakilireward} is provided in Appendix~\ref{sec:relatedworks}.
 
 %However, their results rely on specific assumptions about the relationship between kernel eigenvalues and domain size, which limits the generality of their results. Also, the domain partitioning technique, despite its theoretical appeal, is cumbersome to implement and raises practical concerns, such as the justification of discarding samples and using only those within a subdomain. Our algorithm demonstrates order-optimal results for general kernels using simpler approach leveraging statistical independence, and it only requires a sublinear maximum information gain $\Gamma(n)$, which is always the case. }

% \begin{algorithm}\label{alg:exp}
% \caption{Exploration Phase}
% \begin{algorithmic}
% \REQUIRE $\tau$, $k$, $N$

% \FOR{$n=1,2,\cdots, N$}
%     \STATE Initialize 
%     \FOR{$h=H, H-1, \cdots, 1$}
%         \STATE DO NOTHING! : )
%     \ENDFOR
%     \FOR{$h=1, 2, \cdots, H$}
%         \STATE Observe $s_h$
%         \STATE Select $a_h=\pi_h(s_h):=\arg\max_{a}\sigma_h^n(s_h,a)$  \sattar{\# Note the significant change we are making here}
%     \ENDFOR
% \ENDFOR
% \end{algorithmic}
% \end{algorithm}

% In the next section, we obtain a sufficient number of exploration episodes for designing $\epsilon$-optimal policies, under both scenarios: with and without generative models. 
\subsection{Computational Complexity}
The main computational bottleneck is the matrix inversion in kernel ridge regression, which incurs a cost of $\mathcal{O}(n^3)$, leading to a total complexity of $\mathcal{O}(N^4)$ for our algorithms. This is comparable to the complexities in related work, such as~\citet{vakili2024kernelized} and~\citet{qiu2021reward}.
Notably, the $\mathcal{O}(n^3)$ cost of matrix inversion is not unique to RL but is common across kernel-based supervised learning and bandit literature. Sparse approximation methods, such as Sparse Variational Gaussian Processes (SVGP) and the Nyström method, can significantly reduce this complexity (in some cases, to linear time) while preserving kernel-based confidence intervals and corresponding rates ~\citep[e.g.,][]{vakili2022improved}. However, since these methods are broadly applicable rather than specific to our setting, we chose to maintain a clear, notation-light presentation focused on our main contributions.
\section{ANALYSIS OF THE SAMPLE COMPLEXITY}\label{sec:anal}

In this section, we present our main results on the sample complexity of the algorithms. We first establish a novel confidence interval that is applicable to the unbiased samples collected by our exploration algorithms. We then provide theorems detailing the performance of these algorithms.

\subsection{Confidence Intervals}
We introduce a novel confidence interval that is tighter than existing ones in our RL setting and can also be applied to other RL problems such as offline RL and infinite-horizon settings.
\begin{theorem}[Confidence Bounds]\label{the:conf}

Consider compact sets $\Sc\subset\Rr^{d_s}, \Ac\subset\Rr^{d_a}$, and define $\Zc=\Sc\times \Ac$, $d=d_a+d_s$. Consider two Mercer kernels $k_{\varphi}:\Zc\times\Zc\rightarrow \Rr$ and  $k_{\psi}:\Sc\times\Sc\rightarrow \Rr$. Assume that functions $f:\Zc\rightarrow\Rr$ and $V:\Sc\rightarrow\Rr$, and for each $z\in\Zc$, a conditional probability distribution $P(\cdot|z)$ over $\Sc$, are given such that $f(z)=\E_{s\sim P(\cdot|z)}[V(s)]$, $\|f\|_{\Hc_{k_\varphi}}\le B_1$, $\|V\|_{\Hc_{k_\psi}}\le B_2$, and $\max_{s\in\Sc}V(s)\le v_{\max}$, for some $B_1,B_2, v_{\max}>0$.
Assume a dataset of $\{z_{i}, s'_i\}_{i=1}^n$ is provided, where each $z_i$ is independent of the set $\{s'_j\}_{j=1}^n$, and $s'_i\sim P(\cdot|z_i)$.
Let $\hat{f}^n$ and $\sigma^n$ be the kernel ridge predictor and uncertainty estimator of $f$ using the observations:
\begin{align}\nn
    \hat{f}_n(z) &= k^{\top}_{\varphi_n}(z)(\tau^2I+K_{\varphi_n})^{-1}\bm{y}_{n},\\
    \sigma_n^2(z) &= k_{\varphi}(z,z) - k^{\top}_{\varphi_n}(z)(\tau^2I+K_{\varphi_n})^{-1}k_{\varphi_n}(z),
\end{align}
where $\bm{y}_n=[V(s'_1), V(s'_2), \cdots, V(s'_n))]^{\top}$.
In addition, let $\lambda_m$, $m=1,2,\cdots$ represent the Mercer eigenvalues of $k_{\psi}$ in a decreasing order, and $\psi_m$ the corresponding Mercer eigenfunctions. Assume $|\psi_m|\le \psi_{\max}$ for some $\psi_{\max}>0$. Fix $M\in \Nn$, and let $C$ be a constant such that $C \geq \sum_{m=1}^{M} \lambda_m$. %that serves as an upper bound for the sum of the first $M$ eigenvalues.

Then, for a fixed $z\in\Zc$, and for all $V$, with $\|V\|_{\Hc_{k_\psi}}\le B_2$, we have, the following each hold, with probability at least $1-\delta$,
\begin{equation*}
|f(z) - \hat{f}_n(z)| \le \beta(\delta) \sigma_n(z)
\end{equation*}
%\begin{equation*}
%    f(z) - \hat{f}_n(z) \le \beta(\delta) \sigma_n(z)~~ \text{and} ~~ f(z) - \hat{f}_n(z) \ge -\beta(\delta) \sigma_n(z).
%\end{equation*}
% \begin{equation*}
% \left\{
% \begin{aligned}
% & f(z) - \hat{f}_n(z) \le \beta(\delta) \sigma_n(z),  \\
% & f(z) - \hat{f}_n(z) \ge -\beta(\delta) \sigma_n(z),
% \end{aligned}
% \right.
% \end{equation*}
with $\beta(\delta) =$
\small{\begin{align*}
     B_1+ \frac{C B_2 \psi_{\max} }{\tau}\sqrt{2\log(\frac{M}{\delta})} 
    + \frac{2B_2\psi_{\max}}{\tau}\sqrt{n\sum_{m=M+1}^{\infty}\lambda_m}~.
\end{align*}}

%\aya{, where $C$ is a constant} %that serves as an upper bound for the sum of the first $M$ eigenvalues.}
\end{theorem}

Theorem~\ref{the:conf} provides a confidence bound for kernel ridge regression that is applicable to our RL setting, and is a key result in deriving our sample complexities. 

\paragraph{Proof sketch}
To derive our confidence bounds, we use the Mercer representation of \( V \) and decompose the prediction error \( f(z) - \hat{f}_n(z) \) into error terms corresponding to each Mercer eigenfunction \( \psi_m \). We then divide these terms into two groups: the first \( M \) elements, corresponding to eigenfunctions with the largest eigenvalues, and the remainder. For the top \( M \) eigenfunctions, we establish high-probability bounds using standard kernel-based confidence intervals from \cite{vakili2021optimal}. The remaining terms are bounded based on eigendecay, and we sum over all \( m \) to obtain \( \beta(\delta) \).
%To derive our confidence bounds, we use a novel approach by leveraging the Mercer representation of \( V \) and decomposing the prediction error \( f(z) - \hat{f}_n(z) \) into error terms corresponding to each Mercer eigenfunction \( \psi_m \). We then divide these terms into two groups: the first \( M \) elements, corresponding to eigenfunctions with the largest eigenvalues, and the remainder. For the top \( M \) eigenfunctions, we establish high-probability bounds using standard kernel-based confidence intervals from \citep{vakili2021optimal}. The remaining terms are bounded based on eigendecay, and we sum over all \( m \) to obtain \( \beta(\delta) \).}
\begin{remark}
    Under some mild conditions, for example, the polynomial eigendecay given in Definition~\ref{def:eigendecay}, the following expression can be derived for $\beta$:
    \begin{equation}
        \beta(\delta)= \Oc\left(B_1+\frac{ B_2 \psi_{\max} }{\tau}\sqrt{\log(\frac{n}{\delta})}\right).
    \end{equation}
\end{remark}

With polynomial eigendecay, the remark follows from setting $M$ to $\lceil n^{\frac{1}{p-1}}\rceil 
$ in the expression of $\beta$ in Theorem~\ref{the:conf}.

The confidence interval presented in Theorem~\ref{the:conf} is applicable to a fixed $z\in\Zc$. Over a discrete domain this can be easily extended to all $z\in\Zc$ using a probability union bound and replacing $\delta$ with~$\frac{\delta}{|\Zc|}$ in the expression of $\beta(\delta)$. Using standard discretization techniques, we can also prove a variation of the confidence interval that holds true uniformly over continuous domains. In particular, under the following assumption, we present a variation of the theorem over continuous domains.

\begin{assumption}\label{ass:disc}
For each $n\in\Nn$, there exists a discretization $\Zz$ of $\Zc$ such that, for any $f\in \Hc_k$ with $\|f\|_{\Hc_k}\le B_1$, we have $f(z) - f([z])\le \frac{1}{n}$, where $[z] = {\arg}{\min}_{ z'\in \Zz}||z'-z||_{l^2}$ is the closest point in $\Zz$ to $z$, and $|\Zz|\le cB_1^dn^{d}$, where $c$ is a constant independent of $n$ and $B_1$.
\end{assumption}
Assumption~\ref{ass:disc} is a mild technical assumption that holds for typical kernels~\citep{srinivas2009gaussian, chowdhury2017kernelized, vakili2021optimal}.

\begin{corollary}\label{Cor:cont}
Under the setting of Theorem~\ref{the:conf}, and under Assumption~\ref{ass:disc}, the following inequalities each hold uniformly in $z\in\Zc$ and $V: \|V\|_{\Hc_{k_\psi}}\le B_2$, with probability at least $1-\delta$
\begin{align*}
    f(z)  \le \hat{f}_n(z) + \frac{2}{n} +  \tilde{\beta}(\delta) (\sigma_n(z)+\frac{2}{\sqrt{n}}), \\
    f(z)  \ge\hat{f}_n(z) -\frac{2}{n} -\tilde{\beta}(\delta) (\sigma_n(z)+\frac{2}{\sqrt{n}}),
\end{align*}
% \begin{equation*}
% \left\{
% \begin{aligned}
% & f(z)  \le \hat{f}_n(z) + \frac{2}{n} +  \tilde{\beta}(\delta) (\sigma_n(z)+\frac{2}{\sqrt{n}}),  \\
% & f(z)  \ge\hat{f}_n(z) -\frac{2}{n} -\tilde{\beta}(\delta) (\sigma_n(z)+\frac{2}{\sqrt{n}}),
% \end{aligned}
% \right.
% \end{equation*}
with $\tilde{\beta}(\delta)=\beta(\frac{\delta}{2c_n})$, $c_n=c(u_n(\frac{\delta}{2}))^dn^d$, and $u_n(\delta) = \Oc(\sqrt{n+\log(\frac{1}{\delta}}))$.
%is a $1-\delta$ upper confidence bound on $\|\hat{f}_n\|_{\Hc_k}$.

\end{corollary}

\begin{remark}
    Under some mild conditions, for example, the polynomial eigendecay given in Definition~\ref{def:eigendecay}, the following expression can be derived for $\tilde{\beta}$:
    \begin{equation}
        \tilde{\beta}(\delta)= \Oc\left(B_1+\frac{C B_2 \psi_{\max}}{\tau}\sqrt{d\log(\frac{n}{\delta})}\right).
    \end{equation}
\end{remark}

\subsection{Sample Complexities}

We have the following theorem on the performance of Algorithm~\ref{alg:exp_gen}. 
The weakest assumption one can pose on the value functions is realizability, which posits that the optimal value functions \(V^{\star}_{h}\) for \(h \in [H]\) lie in the RKHS \(H_{k_\psi}\) for some kernel $k_{\psi}:\Sc\times\Sc\rightarrow \Rr$, or at least are well-approximated by \(H_{k_\psi}\). For stateless MDPs or multi-armed bandits where \(H = 1\), realizability alone suffices for provably efficient algorithms~\citep{srinivas2009gaussian, chowdhury2017kernelized, vakili2021optimal}. But it does not seem to be sufficient when \(H > 1\), and in these settings it is common to make stronger assumptions~\citep{jin2020provably, wang2019optimism, chowdhury2023value}.
Following these works, our main assumption is a closure property for all value functions in the following class:
\begin{align}
    \Vc = \left\{
    s\rightarrow \min\left\{
    H, \max_{a\in\Ac}\left\{
    r(s,a) + \aya{\varphi}^{\top}(s,a)\bm{w} + 
    \right. \right. \right. \nonumber \\
    \left. \left. \left. \beta \sqrt{\varphi^{\top}(s,a)\Sigma^{-1}\varphi(s,a)}
    \right\}
    \right\}
    \right\},
\end{align}

where $0<\beta<\infty$, $\|\bm{w}\|\le\infty$ 

and $\Sigma$ is an $\infty\times \infty$ matrix with $\Sigma \succ \tau^2 I$. 

\begin{assumption}[Optimistic Closure]\label{closure_assumption}
For any $V\in\Vc$, for some positive constant $c_v$, we have $\|V\|_{H_{k_\psi}}\leqslant c_v$.
\end{assumption}

This is the same assumption as Assumption~1 in~\cite{chowdhury2023value} and can be relaxed to value functions $\epsilon$ away from this class as described in Section $4.3$ of~\cite{chowdhury2023value}.
The assumption ensures that the proxy value functions (\(V_{h,n}\)) lie within the RKHS of a suitable kernel \(k_{\psi}\). Notably, the RKHS of widely used kernels, such as Matérn and NT kernels, can uniformly approximate any continuous function over compact subsets of \(\mathbb{R}^d\)~\citep{srinivas2009gaussian}.

We have the following theorem on the sample complexity of the exploration algorithm with a generative model.

\begin{theorem}\label{the:gen}
    Consider the reward-free RL framework described in Section~\ref{sec:pf}. Assume the existence of a generative model in the exploration phase that allows the algorithm to select state-action pairs of its choice at each step. Let $N_0$ be the smallest integer satisfying
    \begin{equation*}
2H\beta(\delta)\sqrt{\frac{2\Gamma(N_0)}{N_0\log(1+1/\tau^2)}} +\frac{4\beta(\delta)H}{\sqrt{N_0}} +\frac{4H}{N_0}\le \epsilon,
\end{equation*}
with $\beta(\delta) =\Oc(\frac{H}{\tau}\sqrt{d\log(\frac{NH}{\delta})})$ with a sufficiently large constant. 
Run Algorithm~\ref{alg:exp_gen} for $N\ge N_0$ episodes to obtain the dataset $\Dc_N$. Then, use the obtained samples to design a policy $\pi$ using Algorithm~\ref{alg:plan} with $\beta(\delta) =\Oc(\frac{H}{\tau}\sqrt{d\log(\frac{NH}{\delta})})$ with a sufficiently large constant. 
Then, under Assumptions~\ref{ass:rkhsnorm}, \ref{ass:disc} and~\ref{closure_assumption}, with probability at least $1-\delta$, $\pi$ is guaranteed to be an $\epsilon$-optimal policy. 
\end{theorem}

The following theorem presents the sample complexity for exploration without generative models.

\begin{theorem}\label{the:main}
    Consider the reward free RL framework described in Section~\ref{sec:pf}. Let $N_0$ be the smallest integer satisfying
    \begin{align}\nn
    &\aya{3}H(H+1)\beta(\delta)\sqrt{\frac{2\Gamma(N_0)}{N_0\log(1+1/\tau^2)}} +\frac{8\beta(\delta)H(H+1)}{\sqrt{N_0}} \\\label{eq:suboptgap}
    &+\aya{\frac{4H(H+1)(\log(N_0)+1)}{N_0}+2H\sqrt{N_0(H+1)\log({\frac{2}{\delta})}}}\le \epsilon
\end{align}
with $\beta(\delta) =\Oc(\frac{H}{\tau}\sqrt{d\log(\frac{NH}{\delta})})$ with a sufficiently large constant.     Run Algorithm~\ref{alg:exp2} for $NH\ge N_0H$ episodes to obtain the dataset $\Dc_N$. Then, use the obtained samples to design a policy $\pi$ using Algorithm~\ref{alg:plan}. 
Then, under Assumptions~\ref{ass:rkhsnorm}, \ref{ass:disc} and~\ref{closure_assumption}, with probability at least $1-\delta$, $\pi$ is guaranteed to be an $\epsilon$-optimal policy. 
\end{theorem}

The proof of theorems are provided in Appendix~\ref{appx:gen} and~\ref{appx:main_sample}. 

The expression of suboptimality gap after $N$ samples, given in~\eqref{eq:suboptgap}, can be simplified as
\begin{equation*}
    \Oc\left(  H^3\sqrt{\frac{\Gamma(N)\log(NH/\delta)}{N}}\right).
\end{equation*} 
 
\begin{figure*}[h]
    \centering
    \begin{subfigure}{0.32\textwidth}
        \centering
        \includegraphics[width=\textwidth]{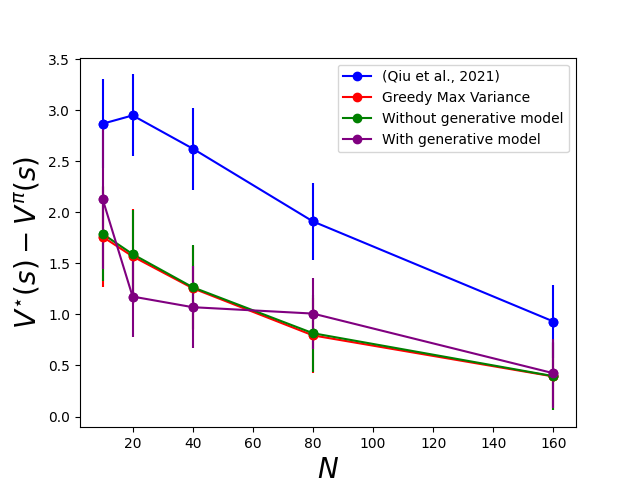}
        \caption{Squared Exponential Kernel}
        \label{fig:RBF_all_algos}
    \end{subfigure}
    %\hspace{0.3em} 
    \begin{subfigure}{0.32\textwidth}
        \centering
        \includegraphics[width=\textwidth]{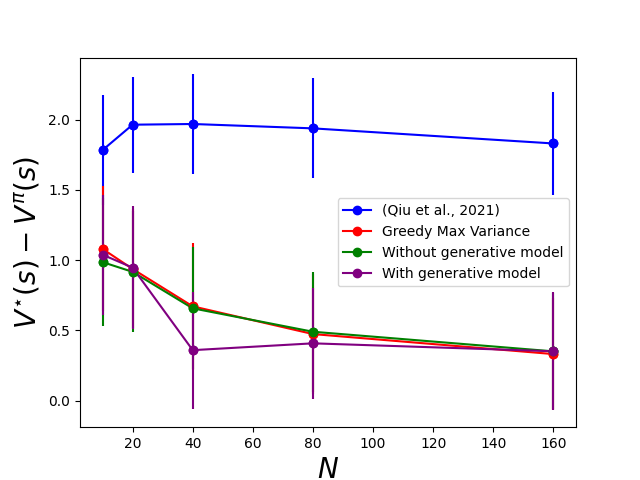} % Replace with the path to your third figure
        \caption{Mat{\'e}rn Kernel with $\nu=2.5$}
        \label{fig:Matern2.5_all_algos}
    \end{subfigure}
    %\hspace{0.3em} 
    \begin{subfigure}{0.32\textwidth}
        \centering
        \includegraphics[width=\textwidth]{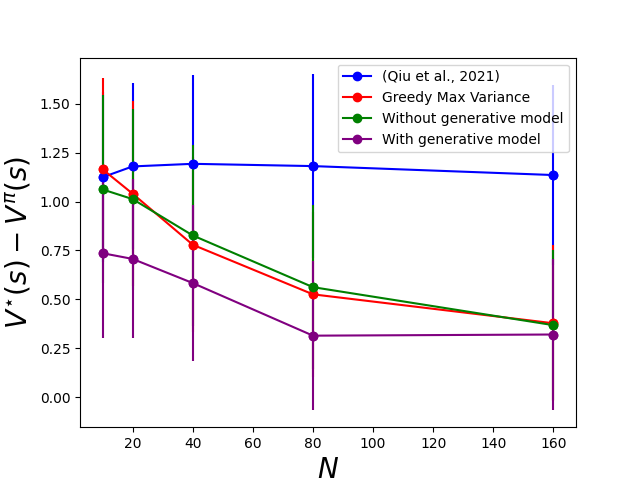} % Replace with the path to your second figure
        \caption{Mat{\'e}rn kernel with $\nu=1.5$}
        \label{fig:Matern1.5_all_algos}
    \end{subfigure}
    \caption{Average suboptimality gap against $N$. The error bars indicate standard deviation.}
    \label{fig:overallresults}
\end{figure*}
\begin{remark}
Replacing $\Gamma(N)=\Oct(N^{\frac{1}{p}})$ in the case of kernels with polynomial eigendecay, we obtain a sample complexity of 
$
    N = \Oct((\frac{H^3}{\epsilon})^{2+\frac{2}{p-1}}).
$
We also recall that without a generative model, we interact with $H$ times more episodes to collect these samples. Specifically, the number of episodes in the exploration phase is 
$NH = \Oct\left(H(\frac{H^3}{\epsilon})^{2+\frac{2}{p-1}}\right)$.

\end{remark}

When specialized for the case of Mat{\'e}rn kernels with $p=1+\frac{2\nu}{d}$, we obtain $NH=\Oct(H(\frac{H^3}{\epsilon})^{2+\frac{d}{\nu}})$ that matches the lower bound for the degenerate case of bandits with $H=1$ proven in~\citet{scarlett2017lower}. Our sample complexity is thus order optimal in terms of $\epsilon$ dependency. We also recall that the existing results lead to possibly vacuous (infinite) sample complexities for these kernels.

\section{EXPERIMENTS}\label{sec:exp}

%    \includegraphics[width=0.32\textwidth]{figures/Matern2.5_all_algos.png} % Use the same path as before
%    \caption{Average suboptimality gap against $N$ for the Mat{\'e}rn Kernel with $\nu=2.5$. The error bars indicate standard deviation.}
%    \label{fig:Matern2.5_all_algos_single}
%\end{figure}

We numerically validate our proposed algorithms and compare with the baseline algorithms. From the literature, we implement~\cite{qiu2021reward}, in which the exploration aims at maximizing a hypothetical reward of $\beta\sigma_n/H$ over each episode $n$. The planning phase is similar to Algorithm~\ref{alg:plan}, but with upper confidence bounds based on kernel ridge regression being used rather than the prediction~$\hat{g}_h$. 
%We numerically validate and compare the performance of several algorithms. We implement the existing approach of~\cite{qiu2021reward} where during the exploration phase at each episode $n$, a policy is designed to obtain high value with respect to a hypothetical reward of $\beta\sigma_n/H$. The planning phase is similar to Algorithm~\ref{alg:plan} with a minor difference that rather than kernel ridge predictions, upper confidence bounds based on kernel ridge regression are used. 
We also implement our exploration algorithms with and without a generative model: Algorithms~\ref{alg:exp_gen} and~\ref{alg:exp2} respectively. Additionally, we implement a heuristic variation of Algorithm~\ref{alg:exp2}, which collects the exploration samples in a greedy manner $a_{h,n}=\argmax_{a\in\Ac}\sigma_{h,n}(s_{h,n},a)$ while remaining on the Markovian trajectory by sampling $s_{h+1}\sim P_h(\cdot|s_h,a_h)$. We refer to this heuristic as \emph{Greedy Max Variance}. For all these  algorithms, we use Algorithm~\ref{alg:plan} to obtain a planning policy. In the experimental setting, we choose $H=10$ and $\Sc=\Ac=[0,1]$ consisting of $100$ evenly spaced points. We choose $r$ and $P$ from the RKHS of a fixed kernel. For the detailed framework and hyperparameters, please refer to Appendix~\ref{appx:exp}. We run the experiment for three different kernels across all $4$ algorithms for $80$ independent runs, and plot the average suboptimality gap $V_1^{\star}(s)-V_1^{\pi}(s)$ for $N=10,20, 40, 80, 160$, as shown in Figure~\ref{fig:overallresults}.
Our proposed Algorithm~\ref{alg:exp2}, without generative model, demonstrates better performance compared to prior work \citep{qiu2021reward} across all three kernels, validating the improved sample efficiency. Notably, \citet{qiu2021reward} performs poorly with nonsmooth kernels. Greedy Max Variance is a heuristic that in many of our experiments performs close to Algorithm~\ref{alg:exp2}.
Furthermore, with access to a generative model, Algorithm~\ref{alg:exp_gen} performs the best. This is anticipated, as the generative model provides the flexibility to select the most informative state-action pairs, unconstrained by Markovian transitions.

\section{CONCLUSION}
\label{sec:conclusion}

We proposed novel algorithms for the kernel-based reward-free RL problem, both with and without generative models. We demonstrated that, with a generative model, a simple algorithm can achieve near-optimal sample complexities. Without the generative model requirement, we showed that an online algorithm achieves the same sample complexity up to a factor of $H$, indicating that the cost of online sampling is a factor of $H$. Our results apply to a general class of kernels, including those with polynomial eigendecay, where existing methods may either lead to vacuous sample complexities~\citep{qiu2021reward} or require additional assumptions and a sophisticated, difficult-to-implement domain partitioning method~\citep{vakilireward}. Our experimental results support these analytical findings. In comparison to the lower bounds proven for the degenerate case of bandits with $H=1$ for the Matérn kernel, the order optimality of our results in terms of $\epsilon$ becomes clear.

% In summary, the analysis of reward-free RL has predominantly focused on tabular and linear settings. However, this paper delves into the kernel-based setting, where prior works have shown suboptimal sample complexity, particularly for non-smooth kernels, or have relied on intricate domain partitioning techniques to address this limitation. Our contribution lies in the introduction of algorithms tailored for both exploration and planning phases, with and without sampling access to a generative model. We demonstrate order-optimal sample complexities in both scenarios, applicable to a wider range of kernels, thus pushing the boundaries of current research. Unlike previous kernel-based methodologies, our algorithm is straightforward and applicable to all Mercer kernels without the need for domain partitioning. Furthermore, we support our theoretical findings with empirical studies that confirm the enhanced sample efficiency compared to other benchmark methods.

% \paragraph{Limitations:}

\clearpage
\bibliography{references}

\begin{thebibliography}{}

\bibitem[Abbasi-Yadkori, 2013]{abbasi2013online}
Abbasi-Yadkori, Y. (2013).
\newblock Online learning for linearly parametrized control problems.

\bibitem[Agarwal et~al., 2020]{agarwal2020model}
Agarwal, A., Kakade, S., and Yang, L.~F. (2020).
\newblock Model-based reinforcement learning with a generative model is minimax optimal.
\newblock In {\em Conference on Learning Theory}, pages 67--83. PMLR.

\bibitem[Antos et~al., 2008]{antos2008learning}
Antos, A., Szepesv{\'a}ri, C., and Munos, R. (2008).
\newblock Learning near-optimal policies with bellman-residual minimization based fitted policy iteration and a single sample path.
\newblock {\em Machine Learning}, 71:89--129.

\bibitem[Auer et~al., 2008]{auer2008near}
Auer, P., Jaksch, T., and Ortner, R. (2008).
\newblock Near-optimal regret bounds for reinforcement learning.
\newblock {\em Advances in Neural Information Processing Systems}, 21.

\bibitem[Azar et~al., 2013]{gheshlaghi2013minimax}
Azar, M.~G., Munos, R., and Kappen, H.~J. (2013).
\newblock Minimax pac bounds on the sample complexity of reinforcement learning with a generative model.
\newblock {\em Machine learning}, 91:325--349.

\bibitem[Bartlett and Tewari, 2012]{bartlett2012regal}
Bartlett, P.~L. and Tewari, A. (2012).
\newblock Regal: A regularization based algorithm for reinforcement learning in weakly communicating {MDPs}.
\newblock {\em arXiv preprint arXiv:1205.2661}.

\bibitem[Bellemare et~al., 2016]{bellemare2016unifying}
Bellemare, M., Srinivasan, S., Ostrovski, G., Schaul, T., Saxton, D., and Munos, R. (2016).
\newblock Unifying count-based exploration and intrinsic motivation.
\newblock {\em Advances in Neural Information Processing Systems}, 29.

\bibitem[Chen and Jiang, 2019]{chen2019information}
Chen, J. and Jiang, N. (2019).
\newblock Information-theoretic considerations in batch reinforcement learning.
\newblock In {\em International Conference on Machine Learning}, pages 1042--1051. PMLR.

\bibitem[Chowdhury and Gopalan, 2017]{chowdhury2017kernelized}
Chowdhury, S.~R. and Gopalan, A. (2017).
\newblock On kernelized multi-armed bandits.
\newblock In {\em International Conference on Machine Learning}, pages 844--853. PMLR.

\bibitem[Chowdhury and Gopalan, 2019]{chowdhury2019online}
Chowdhury, S.~R. and Gopalan, A. (2019).
\newblock Online learning in kernelized {M}arkov decision processes.
\newblock In {\em The 22nd International Conference on Artificial Intelligence and Statistics}, pages 3197--3205. PMLR.

\bibitem[Chowdhury and Oliveira, 2023]{chowdhury2023value}
Chowdhury, S.~R. and Oliveira, R. (2023).
\newblock Value function approximations via kernel embeddings for no-regret reinforcement learning.
\newblock In {\em Asian Conference on Machine Learning}, pages 249--264. PMLR.

\bibitem[Christmann and Steinwart, 2008]{Christmann2008}
Christmann, A. and Steinwart, I. (2008).
\newblock {\em Support Vector Machines}.
\newblock Springer New York, NY.

\bibitem[Domingues et~al., 2021]{domingues2021kernel}
Domingues, O.~D., M{\'e}nard, P., Pirotta, M., Kaufmann, E., and Valko, M. (2021).
\newblock Kernel-based reinforcement learning: A finite-time analysis.
\newblock In {\em International Conference on Machine Learning}, pages 2783--2792. PMLR.

\bibitem[Hazan et~al., 2019]{hazan2019provably}
Hazan, E., Kakade, S., Singh, K., and Van~Soest, A. (2019).
\newblock Provably efficient maximum entropy exploration.
\newblock In {\em International Conference on Machine Learning}, pages 2681--2691. PMLR.

\bibitem[Hu et~al., 2022]{hu2022towards}
Hu, P., Chen, Y., and Huang, L. (2022).
\newblock Towards minimax optimal reward-free reinforcement learning in linear mdps.
\newblock In {\em The Eleventh International Conference on Learning Representations}.

\bibitem[Jacot et~al., 2018]{jacot2018neural}
Jacot, A., Gabriel, F., and Hongler, C. (2018).
\newblock Neural tangent kernel: Convergence and generalization in neural networks.
\newblock {\em Advances in neural information processing systems}, 31.

\bibitem[Jin et~al., 2018]{jin2018q}
Jin, C., Allen-Zhu, Z., Bubeck, S., and Jordan, M.~I. (2018).
\newblock Is {Q}-learning provably efficient?
\newblock {\em Advances in Neural Information Processing Systems}, 31.

\bibitem[Jin et~al., 2020a]{jin2020reward}
Jin, C., Krishnamurthy, A., Simchowitz, M., and Yu, T. (2020a).
\newblock Reward-free exploration for reinforcement learning.
\newblock In {\em International Conference on Machine Learning}, pages 4870--4879. PMLR.

\bibitem[Jin et~al., 2020b]{jin2020provably}
Jin, C., Yang, Z., Wang, Z., and Jordan, M.~I. (2020b).
\newblock Provably efficient reinforcement learning with linear function approximation.
\newblock In {\em Conference on Learning Theory}, pages 2137--2143. PMLR.

\bibitem[Kakade, 2003]{kakade2003sample}
Kakade, S.~M. (2003).
\newblock {\em On the sample complexity of reinforcement learning}.
\newblock University of London, University College London (United Kingdom).

\bibitem[Kearns and Singh, 1998a]{kearns1998}
Kearns, M. and Singh, S. (1998a).
\newblock Finite-sample convergence rates for q-learning and indirect algorithms.
\newblock In {\em Advances in Neural Information Processing Systems}, volume~11. MIT Press.

\bibitem[Kearns and Singh, 1998b]{kearns1998finite}
Kearns, M. and Singh, S. (1998b).
\newblock Finite-sample convergence rates for {Q}-learning and indirect algorithms.
\newblock {\em Advances in Neural Information Processing Systems}, 11.

\bibitem[Lattimore, 2023]{lattimore2023lower}
Lattimore, T. (2023).
\newblock A lower bound for linear and kernel regression with adaptive covariates.
\newblock In {\em The Thirty Sixth Annual Conference on Learning Theory}, pages 2095--2113. PMLR.

\bibitem[Lee and Oh, 2023]{lee2023demystifying}
Lee, J. and Oh, M.-h. (2023).
\newblock Demystifying linear mdps and novel dynamics aggregation framework.
\newblock In {\em The Twelfth International Conference on Learning Representations}.

\bibitem[Levine et~al., 2020]{levine2020offline}
Levine, S., Kumar, A., Tucker, G., and Fu, J. (2020).
\newblock Offline reinforcement learning: Tutorial, review, and perspectives on open problems.
\newblock {\em ArXiv}, abs/2005.01643.

\bibitem[Mercer, 1909]{Mercer1909}
Mercer, J. (1909).
\newblock Functions of positive and negative type, and their connection with the theory of integral equations.
\newblock {\em Philosophical Transactions of the Royal Society of London. Series A, Containing Papers of a Mathematical or Physical Character}, 209:415--446.

\bibitem[Munos and Szepesv{\'a}ri, 2008]{munos2008finite}
Munos, R. and Szepesv{\'a}ri, C. (2008).
\newblock Finite-time bounds for fitted value iteration.
\newblock {\em Journal of Machine Learning Research}, 9(5).

\bibitem[Neu and Pike-Burke, 2020]{neu2020unifying}
Neu, G. and Pike-Burke, C. (2020).
\newblock A unifying view of optimism in episodic reinforcement learning.
\newblock {\em Advances in Neural Information Processing Systems}, 33:1392--1403.

\bibitem[Pathak et~al., 2017]{pathak2017curiosity}
Pathak, D., Agrawal, P., Efros, A.~A., and Darrell, T. (2017).
\newblock Curiosity-driven exploration by self-supervised prediction.
\newblock In {\em International Conference on Machine Learning}, pages 2778--2787. PMLR.

\bibitem[Pedregosa et~al., 2011]{scikit-learn}
Pedregosa, F., Varoquaux, G., Gramfort, A., Michel, V., Thirion, B., Grisel, O., Blondel, M., Prettenhofer, P., Weiss, R., Dubourg, V., Vanderplas, J., Passos, A., Cournapeau, D., Brucher, M., Perrot, M., and Duchesnay, E. (2011).
\newblock Scikit-learn: Machine learning in {P}ython.
\newblock {\em Journal of Machine Learning Research}, 12:2825--2830.

\bibitem[Precup, 2000]{precup2000eligibility}
Precup, D. (2000).
\newblock Eligibility traces for off-policy policy evaluation.
\newblock {\em Computer Science Department Faculty Publication Series}, page~80.

\bibitem[Puterman, 2014]{puterman2014markov}
Puterman, M.~L. (2014).
\newblock {\em Markov decision processes: discrete stochastic dynamic programming}.
\newblock John Wiley \& Sons.

\bibitem[Qiu et~al., 2021]{qiu2021reward}
Qiu, S., Ye, J., Wang, Z., and Yang, Z. (2021).
\newblock On reward-free rl with kernel and neural function approximations: Single-agent {MDP} and {M}arkov {G}ame.
\newblock In {\em International Conference on Machine Learning}, pages 8737--8747. PMLR.

\bibitem[Russo, 2019]{russo2019worst}
Russo, D. (2019).
\newblock Worst-case regret bounds for exploration via randomized value functions.
\newblock {\em Advances in Neural Information Processing Systems}, 32.

\bibitem[Scarlett et~al., 2017]{scarlett2017lower}
Scarlett, J., Bogunovic, I., and Cevher, V. (2017).
\newblock Lower bounds on regret for noisy {G}aussian process bandit optimization.
\newblock In {\em Conference on Learning Theory}, pages 1723--1742. PMLR.

\bibitem[Sidford et~al., 2018a]{sidford2018near}
Sidford, A., Wang, M., Wu, X., Yang, L., and Ye, Y. (2018a).
\newblock Near-optimal time and sample complexities for solving markov decision processes with a generative model.
\newblock {\em Advances in Neural Information Processing Systems}, 31.

\bibitem[Sidford et~al., 2018b]{sidford2018variance}
Sidford, A., Wang, M., Wu, X., and Ye, Y. (2018b).
\newblock Variance reduced value iteration and faster algorithms for solving markov decision processes.
\newblock In {\em Proceedings of the Twenty-Ninth Annual ACM-SIAM Symposium on Discrete Algorithms}, pages 770--787. SIAM.

\bibitem[Srinivas et~al., 2010]{srinivas2009gaussian}
Srinivas, N., Krause, A., Kakade, S.~M., and Seeger, M. (2010).
\newblock Gaussian process optimization in the bandit setting: No regret and experimental design.
\newblock In {\em International Conference on Machine Learning}.

\bibitem[Vakili, 2024]{vakili2024open}
Vakili, S. (2024).
\newblock Open problem: Order optimal regret bounds for kernel-based reinforcement learning.
\newblock In {\em The Thirty Seventh Annual Conference on Learning Theory}, pages 5340--5344. PMLR.

\bibitem[Vakili et~al., 2021a]{vakili2021optimal}
Vakili, S., Bouziani, N., Jalali, S., Bernacchia, A., and Shiu, D.-s. (2021a).
\newblock Optimal order simple regret for {G}aussian process bandits.
\newblock {\em Advances in Neural Information Processing Systems}, 34:21202--21215.

\bibitem[Vakili et~al., 2023]{vakili2023information}
Vakili, S., Bromberg, M., Garcia, J., Shiu, D.-s., and Bernacchia, A. (2023).
\newblock Information gain and uniform generalization bounds for neural kernel models.
\newblock In {\em 2023 IEEE International Symposium on Information Theory (ISIT)}, pages 555--560. IEEE.

\bibitem[Vakili et~al., 2021b]{vakili2021information}
Vakili, S., Khezeli, K., and Picheny, V. (2021b).
\newblock On information gain and regret bounds in gaussian process bandits.
\newblock In {\em International Conference on Artificial Intelligence and Statistics}, pages 82--90. PMLR.

\bibitem[Vakili et~al., 2024]{vakilireward}
Vakili, S., Nabiei, F., Shiu, D.-s., and Bernacchia, A. (2024).
\newblock Reward-free kernel-based reinforcement learning.
\newblock In {\em Forty-first International Conference on Machine Learning}.

\bibitem[Vakili and Olkhovskaya, 2023]{vakili2024kernelized}
Vakili, S. and Olkhovskaya, J. (2023).
\newblock Kernelized reinforcement learning with order optimal regret bounds.
\newblock {\em Advances in Neural Information Processing Systems}, 36.

\bibitem[Vakili et~al., 2022]{vakili2022improved}
Vakili, S., Scarlett, J., Shiu, D.-s., and Bernacchia, A. (2022).
\newblock Improved convergence rates for sparse approximation methods in kernel-based learning.
\newblock In {\em International Conference on Machine Learning}, pages 21960--21983. PMLR.

\bibitem[Wagenmaker et~al., 2022]{wagenmaker2022reward}
Wagenmaker, A.~J., Chen, Y., Simchowitz, M., Du, S., and Jamieson, K. (2022).
\newblock Reward-free rl is no harder than reward-aware rl in linear markov decision processes.
\newblock In {\em International Conference on Machine Learning}, pages 22430--22456. PMLR.

\bibitem[Wang et~al., 2020]{wang2020reward}
Wang, R., Du, S.~S., Yang, L., and Salakhutdinov, R.~R. (2020).
\newblock On reward-free reinforcement learning with linear function approximation.
\newblock {\em Advances in Neural Information Processing Systems}, 33:17816--17826.

\bibitem[Wang et~al., 2019]{wang2019optimism}
Wang, Y., Wang, R., Du, S.~S., and Krishnamurthy, A. (2019).
\newblock Optimism in reinforcement learning with generalized linear function approximation.
\newblock {\em arXiv preprint arXiv:1912.04136}.

\bibitem[Whitehouse et~al., 2023]{whitehouse2024sublinear}
Whitehouse, J., Ramdas, A., and Wu, S.~Z. (2023).
\newblock On the sublinear regret of {GP-UCB}.
\newblock {\em Advances in Neural Information Processing Systems}, 36.

\bibitem[Xie et~al., 2021]{xie2021bellman}
Xie, T., Cheng, C.-A., Jiang, N., Mineiro, P., and Agarwal, A. (2021).
\newblock Bellman-consistent pessimism for offline reinforcement learning.
\newblock {\em Advances in Neural Information Processing Systems}, 34:6683--6694.

\bibitem[Yang and Wang, 2019]{yang2019sample}
Yang, L. and Wang, M. (2019).
\newblock Sample-optimal parametric q-learning using linearly additive features.
\newblock In {\em International conference on machine learning}, pages 6995--7004. PMLR.

\bibitem[Yang and Wang, 2020]{yang2020reinforcement}
Yang, L. and Wang, M. (2020).
\newblock Reinforcement learning in feature space: Matrix bandit, kernels, and regret bound.
\newblock In {\em International Conference on Machine Learning}, pages 10746--10756. PMLR.

\bibitem[Yang et~al., 2020]{yang2020provably}
Yang, Z., Jin, C., Wang, Z., Wang, M., and Jordan, M. (2020).
\newblock Provably efficient reinforcement learning with kernel and neural function approximations.
\newblock {\em Advances in Neural Information Processing Systems}, 33:13903--13916.

\bibitem[Yao et~al., 2014]{yao2014pseudo}
Yao, H., Szepesv{\'a}ri, C., Pires, B.~A., and Zhang, X. (2014).
\newblock Pseudo-{MDPs} and factored linear action models.
\newblock In {\em 2014 IEEE Symposium on Adaptive Dynamic Programming and Reinforcement Learning (ADPRL)}, pages 1--9. IEEE.

\bibitem[Yeh et~al., 2023]{yeh2023sample}
Yeh, S.-Y., Chang, F.-C., Yueh, C.-W., Wu, P.-Y., Bernacchia, A., and Vakili, S. (2023).
\newblock Sample complexity of kernel-based q-learning.
\newblock In {\em International Conference on Artificial Intelligence and Statistics}, pages 453--469. PMLR.

\bibitem[Zanette et~al., 2020]{zanette2020frequentist}
Zanette, A., Brandfonbrener, D., Brunskill, E., Pirotta, M., and Lazaric, A. (2020).
\newblock Frequentist regret bounds for randomized least-squares value iteration.
\newblock In {\em International Conference on Artificial Intelligence and Statistics}, pages 1954--1964. PMLR.

\end{thebibliography}
%%%%%%%%%%%%%%%%%%%%%%%%%%%%%%%%%%%%%%%%%%%%%%%%%%%%%%%%%%%%
\section*{Checklist}

% %%% BEGIN INSTRUCTIONS %%%

 \begin{enumerate}

 \item For all models and algorithms presented, check if you include:
 \begin{enumerate}
   \item A clear description of the mathematical setting, assumptions, algorithm, and/or model. [Yes, we describe the mathematical setting, preliminaries, and problem formulation in Section \ref{sec:pf}. We present the proposed algorithms in Section~\ref{sec:alg} with their corresponding pseudocodes. All theorems and assumptions are stated in Section~\ref{sec:anal}. Theorem~\ref{the:conf} is self-contained, with all necessary assumptions explicitly mentioned in the body of the theorem. It is formulated to be broadly applicable to other problems. Theorems~\ref{the:gen} and~\ref{the:main} rely on Assumptions~\ref{ass:disc} and~\ref{closure_assumption}, which are clearly stated in Section~\ref{sec:anal}.
]
   \item An analysis of the properties and complexity (time, space, sample size) of any algorithm. [Yes, detailed analysis about the confidence intervals and sample complexities of our proposed algorithms are provided in Theorems~\ref{the:conf},~\ref{the:gen} and~\ref{the:main} of Section \ref{sec:anal}. However, we do not provide analysis for time and space complexities.]
   \item (Optional) Anonymized source code, with specification of all dependencies, including external libraries. [Yes, the code used to conduct our experiments is included in a zip file as supplementary material. It also contains a README file and a requirements file to facilitate the installation of all necessary packages.]
 \end{enumerate}

 \item For any theoretical claim, check if you include:
 \begin{enumerate}
   \item Statements of the full set of assumptions of all theoretical results. [Yes, we clearly state the full set of assumptions (Assumptions \ref{ass:disc} and \ref{closure_assumption}) in Section \ref{sec:anal}.] 
   \item Complete proofs of all theoretical results. [Yes, we provide the detailed proofs of Theorems~\ref{the:conf}, ~\ref{the:gen} and~\ref{the:main} in Appendices \ref{appx:conf}, \ref{appx:gen}, \ref{appx:main_sample}, respectively.]
   \item Clear explanations of any assumptions. [Yes, we explicitly state the assumptions of Theorem~\ref{the:conf} within the main body of the theorem, making it self-contained. Assumptions~\ref{ass:disc} and~\ref{closure_assumption}, which are necessary for Theorems~\ref{the:gen} and~\ref{the:main}, are clearly articulated and explained separately.]  
 \end{enumerate}

 \item For all figures and tables that present empirical results, check if you include:
 \begin{enumerate}
   \item The code, data, and instructions needed to reproduce the main experimental results (either in the supplemental material or as a URL). [Yes, we provide the code as an anonymized zip file in the supplementary material, along with a Readme file that instructs the user on how to run the code.]
   \item All the training details (e.g., data splits, hyperparameters, how they were chosen). [Yes, the main paper provides a core explanation of the results, including the
algorithms tested, kernels used, and the synthetic framework in Section \ref{sec:exp}. For comprehensive details (hyperparameter-tuning, visualizations of the reward and transition probability functions), please refer to Appendix \ref{appx:exp}.]
         \item A clear definition of the specific measure or statistics and error bars (e.g., with respect to the random seed after running experiments multiple times). [Yes, our results are accompanied by error bars indicating the standard deviation across 80 independent runs of our experiments.]
         \item A description of the computing infrastructure used. (e.g., type of GPUs, internal cluster, or cloud provider). [Yes, we include in Appendix \ref{appx:exp} the computational resources required for our experiments.]
 \end{enumerate}

 \item If you are using existing assets (e.g., code, data, models) or curating/releasing new assets, check if you include:
 \begin{enumerate}
   \item Citations of the creator If your work uses existing assets. [Yes, we have used the scikit-learn library to implement our kernel-based algorithms, and we have properly cited it (See Appendix \ref{appx:exp}). ]
   \item The license information of the assets, if applicable. [Not applicable]
   \item New assets either in the supplemental material or as a URL, if applicable. [Yes, we submit the code generating our experimental results as a zip file in the supplementary material.]
   \item Information about consent from data providers/curators. [Not Applicable]
   \item Discussion of sensible content if applicable, e.g., personally identifiable information or offensive content. [Not Applicable]
 \end{enumerate}

 \item If you used crowdsourcing or conducted research with human subjects, check if you include:
 \begin{enumerate}
   \item The full text of instructions given to participants and screenshots. [Not Applicable]
   \item Descriptions of potential participant risks, with links to Institutional Review Board (IRB) approvals if applicable. [Not Applicable]
   \item The estimated hourly wage paid to participants and the total amount spent on participant compensation. [Not Applicable]
 \end{enumerate}

 \end{enumerate}
\appendix
% \section{Episodic MDP diagram}

% \begin{figure}[ht]
%     \centering
%     \begin{tikzpicture}

%     \tikzset{state/.style={circle, draw=blue, minimum size=1.1cm, align=center}}
%         % Nodes
%         \node[line width=0.3mm,  blue, state] (s1) {\tcd{$s_1$}};
%         \node[line width=0.3mm,  blue, state, right=of s1] (s2) {\tcd{$s_2$}};
%         \node[right=of s2] (dots) {$\cdots$};
%         \node[line width=0.3mm,  blue, state, right=of dots] (sH) {\tcd{$s_H$}};
%         \node[dashed, line width=0.3mm,  blue, state, right=of sH] (sH1) {\tcd{$s_{H+1}$}};

%         % Edges
%         \draw[line width=0.3mm,  blue, ->] (s1) to[bend left] node[above] {\tcd{$a_1$}} (s2);
%         \draw[line width=0.3mm,  blue,->] (s2) to[bend left] (dots);
%         \draw[line width=0.3mm,  blue,->] (dots) to[bend left] node[above] {\tcd{$a_{H-1}$}} (sH);
%         \draw[dashed, line width=0.3mm,  blue,->] (sH) to[bend left] node[above] {\tcd{$a_{H}$}} (sH1);
%     \end{tikzpicture}
%     \caption{Illustration of an Episodic MDP with an episode of length~$H$.}
%     \label{fig:episodic_mdp}
% \end{figure}

\onecolumn

\section{Related Work}\label{sec:relatedworks}
In this Section, we first provide a technical comparison between our work and that of \citet{qiu2021reward} and \citet{vakilireward}. Following this, we present a more comprehensive literature review, including related works that were not covered in the main paper. We also include a summary table of the sample complexity results in the reward-free RL setting, highlighting our key contributions.

\subsection{Comparison to the Existing Work} \label{appx:comp}

Here, we discuss the key differences between our approach and the closely related works of \citet{qiu2021reward} and \citet{vakilireward}. In ~\citet{qiu2021reward}, they conduct exploration by accumulating standard deviation over an episode, then they apply a planning phase-like algorithm to maximize a reward proportional to $\beta(\delta)\sigma_{h,n}$ at each step of an episode. However, this approach can inflate the confidence interval width multiplier $\beta(\delta)$ by a factor of $\sqrt{\Gamma(n)}$, potentially leading to suboptimal or even trivial sample complexities when $\sqrt{\Gamma(n)}$ is large, as seen in \cite{qiu2021reward}. Specifically, their results are applicable to very smooth kernels like SE, with exponentially decaying Mercer eigenvalues, for which $\Gamma(n)=\Oc(\text{polylog}(n))$. For kernels with polynomial eigendecay, where $\Gamma(n)=\Oc(n^{\frac{1}{p+1}})$ grows polynomially with $n$, this algorithm possibly leads to trivial (infinite) sample complexities. Intuitively, the inflation of $\beta(\delta)$ is due to the  adaptive sampling creating statistical dependencies among observations, specifically through next state transitions. When such dependencies exist, the best existing confidence intervals are based on a kernel adaptation of self-normalized vector values martingales \citep{abbasi2013online}. The $\sqrt{\Gamma(n)}$ term cannot be removed in general for adaptive samples that introduce bias, as was discussed in~\citet{vakili2024open} and~\citet{lattimore2023lower}.  

\cite{vakilireward} utilizes domain partitioning, relying on only a subset of samples to obtain confidence intervals. This approach achieves order-optimal sample complexity for kernels with polynomial eigendecay, offering an $H$-factor improvement compared to our work in the online setting. However, firstly, their results are limited by specific assumptions regarding the relationship between kernel eigenvalues and domain size, which reduces the generality of their findings. Secondly, their domain partitioning method is cumbersome to implement and lacks practical justification, as it requires dropping samples from other subdomains. In contrast, our algorithm achieves order-optimal results for general kernels with a simpler approach that leverages statistical independence. Moreover, our method is well-suited to the generative setting, where their approach offers no clear advantages.

\subsection{Literature Review} \label{appx:lit_review}
\begin{table}[h]
\caption{Existing sample complexities in reward-free RL. $\Sc$, $\Ac$, $H$, $d$ and $p$ represent the state space, action space, episode length, state-action space dimension and parameter of the kernel with polynomial eigendecay, respectively. Last two rows correspond to the performance guarantees for the algorithms proposed in this work.\\ }
\label{samplecomplexitytable}
\centering
\begin{tabular}{ll}
\hline
Setting      & Sample complexity                                                           \\ \hline
Tabular \citep{jin2020reward}     & $\Oc\left(\frac{|\Sc|^2|\Ac| H^5}{\epsilon^2}\right)$                       \\
Linear \citep{wang2020reward}       & $\Oct\left(\frac{d^3 H^6}{\epsilon^2}\right)$                                \\
Kernel-based (exponential eigendecay)  \citep{qiu2021reward} & $\Oc\left(\frac{H^6 \text{polylog} (\frac{1}{\epsilon})}{\epsilon^2}\right)$ \\
Kernel-based (polynomial eigendecay) \citep{vakilireward} 
&$\Oct\left((\frac{H^3}{\epsilon})^{2+\frac{2}{p-1}}\right)$  \\ \hline
\textbf{Kernel-based (exponential eigendecay) (this work)} & $\Oct\left(\frac{H^7 \text{polylog} (\frac{1}{\epsilon})}{\epsilon^2}\right)$ \\ 
\textbf{Kernel-based (polynomial eigendecay) (this work) }
&$\Oct \left(H(\frac{H^3}{\epsilon})^{2+\frac{2}{p-1}}\right)$  \\ \hline
\end{tabular}
\end{table}

Numerous studies have addressed the sample complexity problem in the discounted MDP framework with an infinite horizon, where the agent has sampling access to a generative model, such as \citep{kearns1998finite,gheshlaghi2013minimax,agarwal2020model}. Alternatively, other research has focused on the episodic MDP framework, without reliance on a generative model or an exploratory policy. Both the tabular setting \citep{jin2018q,auer2008near,bartlett2012regal} and the linear setting \citep{jin2020provably,yao2014pseudo,russo2019worst,zanette2020frequentist,neu2020unifying} have been thoroughly examined. Recent literature has extended these techniques to the kernel setting \citep{yang2020provably,yang2020reinforcement,chowdhury2019online,domingues2021kernel,vakili2024kernelized}, although further improvements are needed in achieving better regret bounds. In contrast to these prior works which assume that the reward function is provided, we explore the episodic reward-free setting in this work, both with and without a generative model. This setting is significantly different from standard RL, rendering the existing sample complexity results inapplicable to our context. 

In the context of reward-free RL, numerous empirical studies have proposed various exploration methods from a practical perspective, as demonstrated by works such as \citep{bellemare2016unifying,pathak2017curiosity,hazan2019provably}. Theoretically, researchers have explored the reward-free RL framework across different levels of complexity, ranging from tabular to linear, kernel-based, and deep learning-based models \citep{jin2020reward,wang2020reward, qiu2021reward} (Table \ref{samplecomplexitytable}). %In the tabular case, \citep{jin2020reward} achieve a sample complexity of $\Oc\left(\frac{|\Sc|^2|\Ac| H^5}{\epsilon^2}\right)$, while in the linear function approximation case,  \citep{wang2020reward} prove a sample complexity of $\Oc\left(\frac{d^3 H^6}{\epsilon^2}\right)$. Here $\Sc$, $\Ac$, $H$ and $d$ represent the state and action spaces, the episode length, and state-action space dimension respectively.% 
Although the existing literature adequately covers the tabular and linear settings, it often provides only partial and incomplete findings when addressing the more intricate kernel-based and deep learning settings. The most relevant work in the kernel setting is \citet{qiu2021reward}, which provides a reward-free algorithm whose sample complexity is $\Oc\left(\frac{H^6 \text{polylog} (\frac{1}{\epsilon})}{\epsilon^2}\right)$. Their results however are only applicable to very smooth kernels with exponentially decaying eigenvalues. The recent work of \citet{vakilireward} proved a sample complexity of $\Oct\left((\frac{H^3}{\epsilon})^{2+\frac{2}{p-1}}\right)$ for kernels with polynomial eigendecay. However, they employ a niche domain partitioning technique that, despite its theoretical appeal, is cumbersome to implement and raises practical concerns, as mentioned earlier.
% The recent work by \textcolor{red}{[cite new ICML paper Reward-Free Kernel-Based Reinforcement Learning]} improves over their work by leveraging a domain partitioning technique, inspired by \citep{vakili2024kernelized}, to achieve a better sample complexity applicable to kernels with polynomially decaying eigenvalues. They demonstrate that $\Oc\left((\frac{H^3}{\epsilon})^{2+\frac{2d}{\alpha}}\right)$ exploration episodes are sufficient to obtain $\epsilon$-optimal policies during planning. Here, $\alpha$ denotes the smoothness of the kernel. Our paper contributes by improving the sample complexity within this framework for any positive definite kernel, and by eliminating the need for impractical domain partitioning techniques.

Finally, it's important to mention that the planning phase of our proposed algorithm is similar to the problem of learning a good policy from predefined datasets, typically called batch or offline RL \citep{levine2020offline}. Many prior works on offline RL make the coverage assumption on the dataset, requiring it to sufficiently include any possible state-action pairs with a minimum probability \citep{precup2000eligibility,antos2008learning,chen2019information,munos2008finite}. These works do not address the exploration needed to achieve such good coverage, which is where our reward-free approach significantly differs. Our goal is to demonstrate how to collect sufficient exploration data without any reward information, enabling the design of a near-optimal policy for any reward function during the planning phase.
\section{Proof of Theorem~\ref{the:conf} and Corollary~\ref{Cor:cont}}\label{appx:conf}
For the proof of Theorem~\ref{the:conf}, we leverage the fact that $V$ belongs to an RKHS. Specifically, we use the Mercer representation of $V$
\begin{equation}
    V(s) = \sum_{m=1}^\infty w_m\lambda_m^{\frac{1}{2}}\psi_m(s).
\end{equation}

We can also rewrite the observations in the observation vector $\bm{y}_n$ as the sum of a noise term and the expected value of the observation (noise free part).

\begin{align}
    V(s'_i) = \underbrace{(V(s'_i) - f(z_i))}_{\text{Observation noise}} + \underbrace{f(z_i)}_{\text{Noise-free observation}}
\end{align}

Using the notation $\psibar_m(z)=\E_{s'\sim P(\cdot|z)}\psi_{m}(s')$, we can rewrite $f(z_i)$ as follows

\begin{align}\nn
    f(z_i) &= 
    \E_{s\sim P(\cdot|z_i)}[V(s)]\\\nn
    &= \E_{s\sim P(\cdot|z_i)}\left[\sum_{m=1}^\infty w_m\lambda_m^{\frac{1}{2}}\psi_m(s)\right]\\\nn
    &= \sum_{m=1}^\infty w_m\lambda_m^{\frac{1}{2}}\E_{s\sim P(\cdot|z_i)}[\psi_m(s)]\\
    &= \sum_{m=1}^\infty w_m\lambda_m^{\frac{1}{2}}\psibar_m(z_i)
\end{align}

We then use the following notations, $\varepsilon_i=V(s'_i)-f(z_i)$, $\bm{\varepsilon}_n=[\varepsilon_1, \varepsilon_2, \cdots, \varepsilon_n]^{\top}$, $\bm{f}_n=[f(z_1), f(z_2), \cdots, f(z_n)]^{\top}$, to rewrite the prediction error

\begin{align}\nn
    f(z) -\hat{f}_n(z) &= f(z) - k^{\top}_n(z)(\tau^2I+K_n)^{-1}\bm{y}_n\\\nn
    &= f(z) - k^{\top}_n(z)(\tau^2I+K_n)^{-1}(\bm{\varepsilon}_n+\bm{f}_n)\\\nn
    &= \underbrace{f(z) - k^{\top}_n(z)(\tau^2I+K_n)^{-1}\bm{f}_n}_{\text{Prediction error from noise-free observations}}- \underbrace{k^{\top}_n(z)(\tau^2I+K_n)^{-1}\bm{\varepsilon}_n}_{\text{The error due to noise}}
\end{align}

The first term is deterministic (not random) and can be bounded following the standard approaches in kernel-based models, for example using the following result from \cite{vakili2021optimal}. Let us use the notations 
\begin{equation*}
    \bm{\zeta}_n(z) = k^{\top}_n(z)(\tau^2I+K_n)^{-1}
\end{equation*}
and $\zeta_i(z) =[\bm{\zeta}_n(z)]_i$. 
\begin{lemma}[Proposition~$1$ in \cite{vakili2021optimal}]\label{lem:vakili} We have
\begin{equation*}
    \sigma_n^2(z) = \sup_{f: \|f\|_{\Hc}\le 1} (f(z) - \bm{\zeta}_n^\top(z)\bm{f}_n )^2 + \tau^2\| \bm{\zeta}_n(z)\|^2_{\ell^2}. 
\end{equation*}
\end{lemma}
Based on this lemma, the first term can be deterministically bounded by $B_1\sigma_n(z)$ :

\begin{equation}
    |f(z) - k^{\top}_n(z)(\tau^2I+K_n)^{-1}\bm{f}_n| \le B_1\sigma_n(z)
\end{equation}

We next bound the second term, the error due to noise. 
\begin{align*}
    k^{\top}_n(z)(\tau^2I+K_n)^{-1}\bm{\varepsilon}_n &= \sum_{i=1}^n\zeta_i(z)\varepsilon_i\\
    &=\sum_{i=1}^n \zeta_i(z) (\sum_{m=1}^\infty w_m\lambda_m^{\frac{1}{2}}\psi_m(s'_i) - \sum_{m=1}^\infty w_m\lambda_m^{\frac{1}{2}} \psibar_m(z_i))\\
    &= \sum_{m=1}^\infty w_m\lambda_m^{\frac{1}{2}}\sum_{i=1}^n\zeta_i(z)(\psi_m(s'_i)-\psibar_m(z_i))
    \\
    &= \sum_{m=1}^M w_m\lambda_m^{\frac{1}{2}}\sum_{i=1}^n\zeta_i(z)(\psi_m(s'_i)-\psibar_m(z_i)) + \sum_{m=M+1}^\infty w_m\lambda_m^{\frac{1}{2}}\sum_{i=1}^n\zeta_i(z)(\psi_m(s'_i)-\psibar_m(z_i))
\end{align*}

We note that $\psi_m(s'_i)-\psibar_m(z_i)$ are bounded random variables with a range of $2\psi_{\max}$. Using Chernoff-Hoeffding inequality and the bound on the norm of $\bm{\zeta}_n$ provided in Lemma~\ref{lem:vakili}, we have that with probability at least $1-\delta/M$ 
\begin{equation*}
    \sum_{i=1}^n\zeta_i(z)(\psi_m(s'_i)-\psibar_m(z_i)) \le \frac{\psi_{\max}\sigma_n(z)}{\tau}\sqrt{2\log(\frac{M}{\delta})}.
\end{equation*}

Using a probability union bound, 
with probability $1-\delta$
\begin{align*}
    &\sum_{m=1}^M w_m\lambda_m^{\frac{1}{2}}\sum_{i=1}^n\zeta_i(z)(\psi_m(s'_i)-\psibar_m(z_i)) \\
    &\le \sum_{m=1}^M w_m\lambda_m^{\frac{1}{2}} \frac{\psi_{\max}\sigma_n(z)}{\tau}\sqrt{2\log\left(\frac{M}{\delta}\right)} \\
    &\le \left(\sum_{m=1}^M\lambda_m\right)^{\frac{1}{2}} \left(\sum_{m=1}^M w^2_m\right)^{\frac{1}{2}} \frac{\psi_{\max}\sigma_n(z)}{\tau}\sqrt{2\log\left(\frac{M}{\delta}\right)} \\
    &\le \left(\sum_{m=1}^M\lambda_m\right)^{\frac{1}{2}}B_2\frac{\psi_{\max}\sigma_n(z)}{\tau}\sqrt{2\log\left(\frac{M}{\delta}\right)} \\
    &\le C B_2 \frac{\psi_{\max}\sigma_n(z)}{\tau}\sqrt{2\log\left(\frac{M}{\delta}\right)}
\end{align*}

The second inequality is based on the Cauchy-Schwarz inequality. In the third inequality, we used that $B_2$ is the upper bound on the RKHS norm of $V$. In the last inequality, we used the observation that under both polynomial eigenvalue decay with $p>1$ and exponential eigendecay, the sum of the eigenvalues is bounded by an absolute constant $C$. 
%, independent of $M$.

Also, for the second term, we have
\begin{align*}
\sum_{m=M+1}^\infty w_m\lambda_m^{\frac{1}{2}}\sum_{i=1}^n\zeta_i(z)\left(\psi_m(s'_i)-\psibar_m(z_i)\right)
&\le 
2\psi_{\max}\sum_{m=M+1}^\infty w_m\lambda_m^{\frac{1}{2}}\sum_{i=1}^n\zeta_i(z)
\\
&\le 
2\psi_{\max}\sum_{m=M+1}^\infty w_m\lambda_m^{\frac{1}{2}}\left(n\sum_{i=1}^n\zeta^2_i(z)\right)^{\frac{1}{2}}
\\
&\le \frac{2\sigma_n(z)\psi_{\max}\sqrt{n}}{\tau}\sum_{m=M+1}^\infty w_m\lambda_m^{\frac{1}{2}}\\
&\le \frac{2\sigma_n(z)\psi_{\max}\sqrt{n}}{\tau}\left(\left(\sum_{m=M+1}^\infty w_m^2\right)\left(\sum_{m=M+1}^{\infty}\lambda_m\right)\right)^{\frac{1}{2}}\\
& \le \frac{2B_2\sigma_n(z)\psi_{\max}}{\tau}\left(n\sum_{m=M+1}^{\infty}\lambda_m\right)^{\frac{1}{2}}.
\end{align*}

The first inequality holds by definition of $\psi_{\max}$. The second inequality is based on the Cauchy-Schwarz inequality. The third inequality uses Lemma~\ref{lem:vakili}. The fourth inequality utilizes the Cauchy-Schwarz inequality again, and the last inequality results from the upper bound on the RKHS norm of~$V$.

Putting together, with probability $1-\delta$,
\begin{equation}
    k^{\top}_n(z)(\tau^2I+K_n)^{-1}\bm{\varepsilon}_n \le \frac{ C B_2\psi_{\max}\sigma_n(z)}{\tau}\sqrt{2\log(\frac{M}{\delta})}  + \frac{2B_2\sigma_n(z)\psi_{\max}}{\tau}\sqrt{n\sum_{m=M+1}^{\infty}\lambda_m}.
\end{equation}

\paragraph{Proof of Corollary~\ref{Cor:cont}}
To extend the confidence interval given in Theorem~\ref{the:conf} to hold uniformly on $\Zc$, we use a discretization argument. For this purpose, we apply Assumption~\ref{ass:disc} to $f$ and $\hat{f}_n$, and also use Assumption~\ref{ass:disc} to bound the discrimination error in $\sigma_n$. The following lemma provides a high probability bound on $\|\hat{f}_n\|_{k_{\varphi}}$.

\begin{lemma}\label{lem:rkhshatf}
    For function $f$ defined in Theorem~\ref{the:conf}, the RKHS norm of $\hat{f}_n$ satisfies the following with probability at least $1-\delta$:
    \begin{equation}
    \|\hat{f}_n\|_{\Hc_{k_\varphi}} \le B_1 +\frac{v_{\max}}{\tau}\sqrt{2 \aya{(\Gamma_{k_{\varphi}}(n)+1+\log(\frac{1}{\delta}))}}.
    \end{equation}
\end{lemma}

For a proof see Lemma~5 in~\citep{vakili2024kernelized}.

Let $B_3(\delta) = B_1 +\frac{v_{\max}}{\tau}\sqrt{2(\Gamma_{k_{\varphi}}(n)+1+\log(\frac{1}{\delta}))}$ denote the $1-\delta$ upper confidence bound on $\|\hat{f}_n\|_{H_{k_{\varphi}}}$. Let $\Zz$ be the discretization of $\Zc$ specified in Assumption~\ref{ass:disc} with RKHS norm bound $B_3(\frac{\delta}{2})$. That is for any $g\in \Hc_{k_{\varphi}}$ with $\|g\|_{\Hc_{k_{\varphi}}}\le B_3(\frac{\delta}{2})$, we have $g(z)-g([z])\le \frac{1}{n}$, where $[z]=\argmin_{z'\in\Zz}\|z'-z\|$ is the closest point in $\Zz$ to $z$, and $|\Zz|\le c_n$, where $c_n=c(B_3(\frac{\delta}{2}))^dn^d$. Applying Assumption~\ref{ass:disc} to $f$ and $\hat{f}_n$ with this discretization, it holds for all $z\in \Zc$ that
\begin{equation}\label{eq:mm}
    |f(z)-f([z])|\le \frac{1}{n}. 
\end{equation}
In addition, by Lemma~\ref{lem:rkhshatf}, with probability eat least $1-\delta$
\begin{equation}\label{eq:mm2}
    |\hat{f}_n(z)-\hat{f}_n([z])\le \frac{1}{n}. 
\end{equation}

Furthermore, we have the following lemma, which can roughly be viewed as a Lipschitz continuity property for $\sigma_n$. 

\begin{lemma}\label{lem:sigmadisc}
    Under Assumption~\ref{ass:disc}, with the discrimination $\Zz$ described above, it holds for all $z\in\Zc$ that
    \begin{equation*}
        \sigma_n(z) -\sigma_n([z]) \le \frac{2}{\sqrt{n}}.
    \end{equation*}
\end{lemma}

\begin{proof}[Proof of Lemma~\ref{lem:sigmadisc}]

Using the reproducing property of RKHS, we have 
\begin{align}
    \|k_n^{\top}(z)(K_n+\tau^2I)^{-1}k_n(\cdot)\|_{\Hc_k} \le \frac{k_{\max}\sqrt{n}}{\tau},
\end{align}
where $k_{\max}$ is the maximum value of the kernel. 
Let us define $q(\cdot, \cdot') = k_n^{\top}(\cdot)(K_n+\tau^2I)^{-1}k_n(\cdot')$. We can write 
\begin{align*}\nn
&\vspace{-3em}|\sigma^2_n(z) - \sigma_n^2([z]) |\\
&=\big|(k(z,z)-q(z,z))-(k([z],[z])-q([z],[z]))\big|\\
&=\big|(k(z,z)-q(z,z)) -(k(z,[z])-q(z,[z]))
+(k(z,[z])-q(z,[z]))
-(k([z],[z])-q([z],[z]))\big|\\\nn
&\le |k(z,z)-k(z,[z])| + |k(z,[z])
-k([z],[z])| +|q(z,z)-q(z,[z])| \aya{+} |q(z,[z])
-q([z],[z])|\\\nn
&\le\frac{4}{n}.
\end{align*}

To obtain a discretization error bound for the standard deviation from that of the variance, we write
\begin{align}\nn
(\sigma_n(z) - \sigma([z]))^2&\le
\left|\sigma_n(z) - \sigma([z])\right|(\sigma_n(z) + \sigma([z]))\\\nn
&=|\sigma^2_n(z) - \sigma^2([z])|\\\nn
&\le\frac{4}{n}.
\end{align}

Therefore,
\begin{equation*}
|\sigma_n(z) - \sigma([z])|\le\frac{2}{\sqrt{n}}.
\end{equation*}

\end{proof}

Applying a probability union bound on the discretization $\Zz$ to Theorem~\ref{the:conf}, and considering the error bounds in~\eqref{eq:mm},~\eqref{eq:mm2} and Lemma~\ref{lem:sigmadisc}, we arrive at Corollary~\ref{Cor:cont}.

\section{Proof of Theorem~\ref{the:gen}}\label{appx:gen}

First, we define the following high-probability event :
\begin{equation}\label{eq:event}
    \Ec = \left\{  \forall h\in[H], |\hat{g}_{h}(z)-[P_hV_{{h+1}}](z)|\le \beta(\delta)\left(\sigma_{h,N}(z)+\frac{2}{\sqrt{n}}\right)+\frac{2}{n}\right\},
\end{equation}
where $\beta(\delta)= \Oc\left(\frac{H}{\tau}\sqrt{d\log(\frac{NH}{\delta})}\right)$ as specified in Corollary~\ref{Cor:cont} with $B_1=\Oc(H)$ and $B_2=c_v$. 
Using Corollary~\ref{Cor:cont}, we have $\Pr[\Ec]\ge 1-\delta$.

We divide the rest of the analysis into several steps, as outlined next.

\paragraph{Step 1:} Under $\Ec$, with reward $r$, we bound $V^{\star}_1(s)-V^{\pi}_1(s)$ using $ V_1(s)-V^{\pi}_1(s)$, based on the following lemma. Recall that $V^{\pi}_h$ and $V^\star_h$ are the value functions of policy $\pi$ and the optimal policy, respectively, and $V_h$ is the proxy value functions used in Algorithm~\ref{alg:plan}. 

\begin{lemma}\label{lem:VstarVplan}
    Under $\Ec$, we have
    \begin{equation}
        V^{\star}_h(s)- V_h(s) \le (H+1-h)(\frac{2\beta(\delta)}{\sqrt{N}}+\frac{2}{N}).
    \end{equation}
\end{lemma}    
\begin{proof}[Proof of Lemma~\ref{lem:VstarVplan}]

The lemma is proven by induction over $h$, starting from $V^{\star}_{H+1} = V_{H+1}=\bm{0}$. We have
\begin{align*}
    Q^{\star}_h(s,a) - Q_{h}(s,a)
    &= r_h(s,a)+ [P_hV^{\star}_{h+1}](s,a)- r_h(s,a) - \hat{g}_{h}(s,a)- \beta(\delta)\sigma_{h,N}(s,a)\\
    &\le [P_hV^{\star}_{h+1}](s,a) -[P_hV_{h+1}](s,a)+\frac{2\beta(\delta)}{\sqrt{N}}+\frac{2}{N}\\
    &=[P_h(V^{\star}_{h+1}-V_{h+1})](s,a)+\frac{2\beta(\delta)}{\sqrt{N}}+\frac{2}{N}\\
    &\le (H+1-h)(\frac{2\beta(\delta)}{\sqrt{N}}+\frac{2}{N})
\end{align*}
The first inequality holds by $\Ec$, and the second inequality by induction assumption.
Then, we have
\begin{align*}
    V^{\star}_h(s_h) - V_h(s_h) &=
    \max_{a\in\Ac} Q^{\star}_h(s,a) - \max_{a\in\Ac}Q_{h}(s,a)\\
    &\le \max_{a\in\Ac} \{Q^{\star}_h(s,a) - Q_{h}(s,a)\}\\
    &\le (H+1-h)(\frac{2\beta(\delta)}{\sqrt{N}}+\frac{2}{N}).
\end{align*}
That proves the lemma. 

\end{proof}

\paragraph{Step 2:} We also bound $V_1(s)-V^{\pi}_1(s) $ using the sum of standard deviations for the trajectory generated by the policy. 

\begin{lemma}\label{VHVpisum}
    Under $\Ec$, we have 
    \aya{
    \begin{equation*}
        V_1(s_1)-V^{\pi}_1(s_1)  \le \E\left[\sum_{h=1}^H 2 \beta(\delta)\sigma_{h,N}(s_h,a_h) \right] +\frac{2 H \beta(\delta)}{\sqrt{N}}+\frac{2 H}{N},
    \end{equation*}
    }
where the expectation is taken with respect to the trajectory generated by the policy.  
\end{lemma}

\begin{proof}[Proof of Lemma~\ref{VHVpisum}]
Note that $V_{H+1}=V^{\pi}_{H+1}=\bm{0}$. We next obtain a recursive relationship for the difference $V_h(s)-V^{\pi}_h(s)$. 
\begin{align*}
    V_h(s_h)-V^{\pi}_h(s_h) &= Q_h\left(s_h, \pi(s_h)\right) -  Q^{\pi}_h\left(s_h, \pi(s_h)\right) \\
    &=r\left(s_h, \pi(s_h)\right) + \hat{g}_h\left(s_h, \pi(s_h)\right) + \beta(\delta)\sigma_{h,N}\left(s_h, \pi(s_h)\right) - r\left(s_h, \pi(s_h)\right) - [P_hV^{\pi}_{h+1}]\left(s_h, \pi(s_h)\right)\\
    &\le [P_hV_{h+1}]\left(s_h, \pi(s_h)\right) +2\beta(\delta)\sigma_{h,N}\left(s_h, \pi(s_h)\right) + \frac{2\beta(\delta)}{\sqrt{N}}+\frac{2}{N}- [P_hV^{\pi}_{h+1}]\left(s_h, \pi(s_h)\right),\\
\end{align*}
where the inequality is due to $\Ec$. 
Recursive application of the above inequality over $h=H, H-1, \cdots, 1$, we obtain
\begin{align*}
    V_1(s_1) - V_1^{\pi}(s_1)& \le \E_{s_{h+1}\sim P(\cdot|s_h,\pi(s_h)), h < H}\left[\sum_{h=1}^H2\beta(\delta)\sigma_{h,N}\left(s_h, \pi(s_h)\right)\right]+ \frac{2H\beta(\delta)}{\sqrt{N}}+\frac{2H}{N}.
\end{align*}

\end{proof}

\paragraph{Step 3:} By definition, we have $V_1^{\pi}(s_1; \beta(\delta)\sigma_{N} )\le V_1^{\star}(s_1; \beta(\delta)\sigma_{N})$. Note that $V^{\pi}_1(s_1; \beta(\delta)\sigma_{N})= \beta (\delta)\sum_{h=1}^H\sigma_{h,N}(s_h, \pi(s_h))$.

\paragraph{Step 4:} We have $V^{\star}_1(s;\beta(\delta)\sigma_{N} ) \le V_1^{\star}(s; \beta(\delta)\sigma_{n} )$. This is due to the observation that $\sigma_{h,n}$ is decreasing in the number $n$ of observations. We note that conditioning on observations only reduces the variance. That is seen from the positive definiteness of the Gram matrix and the formula for kernel ridge uncertainty estimator given in~\eqref{eq:krrvar}. 

\paragraph{Step 5:} Recall the selection rule in Algorithm~\ref{alg:exp_gen}: $s_{h,n},a_{h,n} = \argmax_{s,a}\sigma_{h,n-1}(s,a)$.
When exploring with generative model, with this rule of selection, we have $V_1^{\star}(s_1; \beta(\delta)\sigma_{n-1} ) \le \beta(\delta)\sum_{h=1}^H \sigma_{h,n-1}(s_{h,n}, a_{h,n})$. 
\paragraph{Step 6:} Combining all previous steps, we conclude that, under $\Ec$,
\begin{equation}
    V_1^{\star}(s) - V_1^{\pi}(s) \le \frac{2\beta(\delta)}{N}\sum_{n=1}^N\sum_{h=1}^H\sigma_{h,n-1}(s_{h,n},a_{h,n})+\frac{4\beta(\delta)H}{\sqrt{N}} +\frac{4H}{N}.
\end{equation}

\paragraph{Step 7:}

We bound the sum of standard deviations according to the following lemma that is a kernel based version of elliptical potential lemma~\citep{abbasi2013online}. 
\begin{lemma}
    For each $h$, we have
    \begin{equation}
        \sum_{n=1}^N\sigma^2_{h,n-1}(s_{h,n},a_{h,n})\le \frac{2\Gamma(N)}{\log(1+1/\tau^2)}.
    \end{equation}
\end{lemma}
See, e.g., \cite{srinivas2009gaussian} for a proof. Using Cauchy–Schwarz inequality, we obtain
\begin{equation*}
    \sum_{n=1}^N\sigma_{h,n-1}(s_{h,n},a_{h,n}) \le \sqrt{\frac{2N\Gamma(N)}{\log(1+1/\tau^2)}}.
\end{equation*}

\paragraph{Step 8:} From Steps $6$ and $7$, we conclude that, $\pi$ is an $\epsilon$-optimal policy with $\epsilon$ no larger than 
\begin{equation*}
    V_1^{\star}(s) - V_1^{\pi}(s) \le 
    2H\beta(\delta)\sqrt{\frac{2\Gamma(N)}{N\log(1+1/\tau^2)}} +\frac{4\beta(\delta)H}{\sqrt{N}} +\frac{4H}{N}.
\end{equation*}

A simpler expression can be given as

\begin{equation*}
    V_1^{\star}(s) - V_1^{\pi}(s) =\Oc\left(  H^2\sqrt{\frac{\Gamma(N)\log(NH/\delta)}{N}}\right).
\end{equation*}

Now, let $N_0$ be the smallest integer such that the right hand side less than $\epsilon$. For any $N\ge N_0$ the suboptimality gap of the policy is at most $\epsilon$.
This completes the proof of Theorem~\ref{the:gen}. 

\section{Proof of Theorem~\ref{the:main}}\label{appx:main_sample}

We define the event $\Ec$ similar to the proof of Theorem~\ref{the:gen}. The first $4$ steps related to the planning phase are exactly the same as in the proof of Theorem~\ref{the:gen}. The rest of the proof is different and we will present it here.

In addition to $\Ec$, we define another high-probability event $\Ec'$ where all the confidence intervals utilized in the exploration hold true. Specifically, we define the following:
\begin{equation}\label{eq:event2}
    \Ec' = \{\forall n\in[N], \forall h\in[H], |\hat{f}_{h,n}(z)-f_{h,n}(z)|\le \beta(\delta)(\sigma_{h,n}(z)+\frac{2}{\sqrt{n}})+\frac{2}{n}\},
\end{equation}
where \aya{$\beta(\delta)= \Oc\left(\frac{H}{\tau}\sqrt{d\log(\frac{NH}{\delta})}\right)$. }
Using Corollary~\ref{Cor:cont}, we have $\Pr[\Ec']\ge 1-\delta$.

The following steps are specific to the exploration without the generative model. We define a reward sequence using $\tilde{\sigma}^{h_0}_{h, n}$ such that $\tilde{\sigma}^{h_0}_{h, n}=\sigma_{h,n}$ when $h=h_0$ and $\tilde{\sigma}^{h_0}_{h, n}=0$ for all $h\neq h_0$. 

\paragraph{Step 5b:}
We have
$V^{\star}_1(s; \beta(\delta)\sigma_{n})\le \sum_{h_0=1}^H V^{\star}_1(s; \beta(\delta)\tilde{\sigma}^{h_0}_{ n})$. Note that the optimal policy with rewards $\tilde{\sigma}^{h_0}_{h, n}$ optimizes $\sigma_{h,n}$ at step $h=h_0$, while the optimal policy with rewards $\sigma_{n}$ optimizes the sum of $\sigma_{h,n}$ over all steps. 

\paragraph{Step 6b:} We bound $V^{\star}_1(s; \beta(\delta)\tilde{\sigma}^{h_0}_{ n})$ using  $ V_{1,(n,h_0)}(s)$ where $V_{h,(n,h_0)}$ is the proxy for the value function used in Algorithm~\ref{alg:exp2}. 

\begin{lemma}\label{lem:vnstar_vn}
  Under event $\Ec'$, for all $s\in\Sc$, 
    \begin{equation*}
        V^{\star}_h(s; \beta(\delta)\tilde{\sigma}^{h_0}_{ n})\le V_{h,(n,h_0)}(s)+(h_0+1-h)(\frac{2\beta(\delta)}{\sqrt{n}}+\frac{2}{n}).
    \end{equation*}
\end{lemma}

\begin{proof}[Proof of Lemma~\ref{lem:vnstar_vn}]

    The lemma is proven by induction, starting from $V^{\star}_{h_0+1}(\cdot; \beta(\delta) \tilde{\sigma}^{h_0}_{ n})=V_{h_0+1,(n,h_0)}=\bm{0}$. We have, for $h\le h_0$
    \begin{align}\nn
        V^{\star}_{h}(s;  \beta(\delta)\tilde{\sigma}^{h_0}_{ n})-V_{h,(n,h_0)}(s) &= \max_{a\in\Ac} Q^{\star}_{h}(s,a;  \beta(\delta)\tilde{\sigma}^{h_0}_{ n})-\max_{a\in\Ac}Q_{h,(n,h_0)}(s,a)\\\nn
        &\le \max_{a\in\Ac} \left\{Q^{\star}_{h}(s,a; \beta(\delta) \tilde{\sigma}^{h_0}_{ n})-Q_{h,(n,h_0)}(s,a)\right\}\\\nn
        & \le\max_{a\in\Ac} \left\{[P_hV^{\star}_{h+1}](s,a;  \beta(\delta)\tilde{\sigma}^{h_0}_{ n})-[P_hV_{h+1,n}](s,a)+\frac{2\beta(\delta)}{\sqrt{n}}+\frac{2}{n}\right\}\\\nn
        &\le (h_0+1-h)(\frac{2\beta(\delta)}{\sqrt{n}}+\frac{2}{n}).
    \end{align}
    The first inequality is due to rearrangement of $\max$, the second inequality holds under $\Ec'$, and the third inequality is by the base of induction. We thus prove the lemma. 
\end{proof}

\paragraph{Step 7b:} Fix $n$ and $h_0$.
Let $\pi_{n,h_0}$ denote the exploration policy in episode $nH+h_0$.
We bound $V_{1,(n,h_0)}(s_1)- V^{\pi_{n,h_0}}_1(s_1; \beta(\delta)\tilde{\sigma}^{h_0}_{n-1})
$ using $
2\beta(\delta)\sum_{h=1}^{h_0}\sigma_{h,n-1}(s_h, a_h) $. Here for simplicity of notation, we use $s_h$ and $a_h$ for the state and action at step $h$ of episode corresponding to $n$ and $h_0$. In a richer notation, $n$ and $h_0$ should be specified. 
 
\begin{lemma}\label{lem:v_minus_vpi}
    Under event $\Ec'$, we have 
    \begin{align*}
        &V_{1,(n,h_0)}(s_1)- V^{\pi_{n,h_0}}_1(s_1; \beta(\delta)\tilde{\sigma}^{h_0}_{n-1})\le \sum_{h=1}^{h_0}(2\beta(\delta)\sigma_{h,n-1}(s_h,a_h)+\frac{2\beta(\delta)}{\sqrt{n}}+\frac{2}{n})\\\nn
        &+ \sum_{h=1}^{h_0}([P_hV_{h+1,n}](s_h, a_h) - V_{h+1, n}(s_{h+1}))
    + \sum_{h=1}^{h_0}(V^{\pi_{n,h_0}}_{h+1}(s_{h+1}; \beta(\delta)\tilde{\sigma}^{h_0}_{n-1}) - [P_hV^{\pi_{n,h_0}}_{h+1}](s_h, a_h; \beta(\delta)\tilde{\sigma}^{h_0}_{n-1})).
    \end{align*}
\end{lemma}

The second and third terms are martingale sums which can be bounded using Azuma-Hoeffding inequality, we refer to them as
\begin{equation*}
    \zeta_{h,(n,h_0)} = [P_hV_{h+1,n}](s_h, a_h) - V_{h+1, n}(s_{h+1})
\end{equation*}

\begin{equation*}
    \xi_{h,(n,h_0)} = V^{\pi_{n,h_0}}_{h+1}(s_{h+1}; \beta(\delta)\tilde{\sigma}^{h_0}_{n-1}) - [P_hV^{\pi_{n,h_0}}_{h+1}](s_h, a_h; \beta(\delta)\tilde{\sigma}^{h_0}_{n-1})
\end{equation*}

\begin{proof}[Proof of Lemma~\ref{lem:v_minus_vpi}]

We obtain a iterative relation over $h$. In particular
\begin{align*}
    V_{h,(n,h_0)}(s_h)- V^{\pi_{n,h_0}}_h(s_h; \beta(\delta)\tilde{\sigma}^{h_0}_{n-1})&=Q_{h,(n,h_0)}(s_h, a_h)- Q^{\pi_{n,h_0}}_h(s_h, a_h; \beta(\delta)\tilde{\sigma}^{h_0}_{n-1})\\
    & \hspace{-7em}\le  [P_hV_{h+1,n}](s_h, a_h) - [P_hV^{\pi_{n,h_0}}_{h+1}](s_h, a_h; \beta(\delta)\tilde{\sigma}^{h_0}_{n-1})
    + 2\beta(\delta)\sigma_{h,n-1}(s_h,a_h)+\frac{2\beta(\delta)}{\sqrt{n}}+\frac{2}{n}\\
    &\hspace{-7em} = V_{h+1,(n,h_0)}(s_h)- V^{\pi_{n,h_0}}_{h+1}(s_h; \beta(\delta)\tilde{\sigma}^{h_0}_{n-1})+ 2\beta(\delta)\sigma_{h,n-1}(s_h,a_h)+\frac{2\beta(\delta)}{\sqrt{n}}+\frac{2}{n} + \zeta_{h,(n,h_0)} + \xi_{h,(n,h_0)}.
\end{align*}

Iterating over $h$ and noticing $V_{h_0+1,(n,h_0)}- V^{\pi_{n,h_0}}_{h_0+1}(\cdot; \beta(\delta)\tilde{\sigma}^{h_0}_{n-1}) =\bm{0}$ the lemma is proven.

\end{proof}

\paragraph{Step 8b:}
We note that $V^{\pi_{n,h_0}}_{1,(n,h_0)}(s)=\beta(\delta)\sigma_{h,n-1}(s_{h_0, (n,h_0)}, a_{h_0,(n,h_0)})$.

\paragraph{Step 9b:}

Combining Steps 5b-8b we conclude that
\begin{equation}
    V_1^{\star}(s\aya{;} \beta(\delta)\sigma_{n-1}) \le \sum_{h_0=1}^H\sum_{h=1}^{h_0}\left(3\beta(\delta)\left(\sigma_{h,n-1}(s_{h,(n,h_0)}, a_{h,(n,h_0)})\right)+\frac{4\beta(\delta)}{\sqrt{n}}+\frac{4}{n} +\zeta_{h,(n,h_0)}+\xi_{h,(n,h_0)}\right).
\end{equation}

\paragraph{Step 10b:} Combining with previous steps similar to the proof of Theorem~\ref{the:gen}, and using Azuma-Hoeffding inequality on $\zeta_{h,(n,h_0)}$ and $ \xi_{h,(n,h_0)}$, we get, with probability $1-\delta$
\begin{equation*}
    V_1^{\star}(s) - V_1^{\pi}(s) \le \aya{3}H(H+1)\beta(\delta)\sqrt{\frac{2\Gamma(N)}{N\log(1+1/\tau^2)}} +\frac{8\beta(\delta) H(H+1)}{\sqrt{N}} + \aya{\frac{4H (H+1) (\log(N)+1)}{N}+2H\sqrt{N(H+1)\log({\frac{2}{\delta})}}}.
\end{equation*}

The expression can be simplified as

\begin{equation*}
    V_1^{\star}(s) - V_1^{\pi}(s) =\Oc\left(  H^3\sqrt{\frac{\Gamma(N)\log(\aya{NH}/\delta)}{N}}\right).
\end{equation*}

Now, let $N_0$ be the smallest integer such that the right hand side less than $\epsilon$. For any $N\ge N_0$ the suboptimality gap of the policy is at most $\epsilon$.
This completes the proof of Theorem~\ref{the:main}.

%\newpage
\section{Experimental Details}
\label{appx:exp}

%For the experimental setting, we choose $H=10$, $\Sc=\Ac=[0,1]$ consisting of $100$ evenly spaced points. The reward function and the conditional probability distribution are chosen as arbitrary functions in the RKHS of a fixed kernel: Squared Exponential (SE), Mat{\'e}rn kernels with parameter $\nu=2.5$ and $\nu=1.5$. For all kernels we use lengthscale of $0.1$. With SE kernel we use $\tau=0.01$, and with Met{\'e}rn kernels we use $\tau=0.5$.
%For fixed reward and transition probability distribution, we repeat the experiment with all $4$ algorithms for $80$ independent runs and plot the average sub-optimality of the algorithms $V_1^{\star}(s)-V^{\pi}(s)$ for values $N=10,20, 40, 80, 160$. We also perform hyperparameter tuning for the confidence interval width multiplier. While the theoretical values for $\beta$, especially in~\cite{qiu2021reward}, tend to be high, we fine-tune  $\beta$ and select the best one ($\beta=0.1$). %run the experiments with various $\beta$ values and pick the one performing the best ($\beta=0.1$ for the experiments shown here).
%Further experimental details, including environment specifications, simulations with different $\beta$ values, and additional experiments, can be found in Appendix~\ref{appx:exp}.

Here, we outline the procedure for generating $r$ and $P$ test functions from the RKHS, the computational resources utilized for the simulations, and the fine-tuning process of the confidence interval width multiplier $\beta$. Additionally, we present further experimental results when various samples of $r$ and $P$ are drawn from the RKHS.
\subsection{Synthetic Test Functions from the RKHS}
\begin{figure}[h]
    \centering
    \begin{subfigure}[b]{0.3\textwidth}
        \centering
        \includegraphics[width=\textwidth]{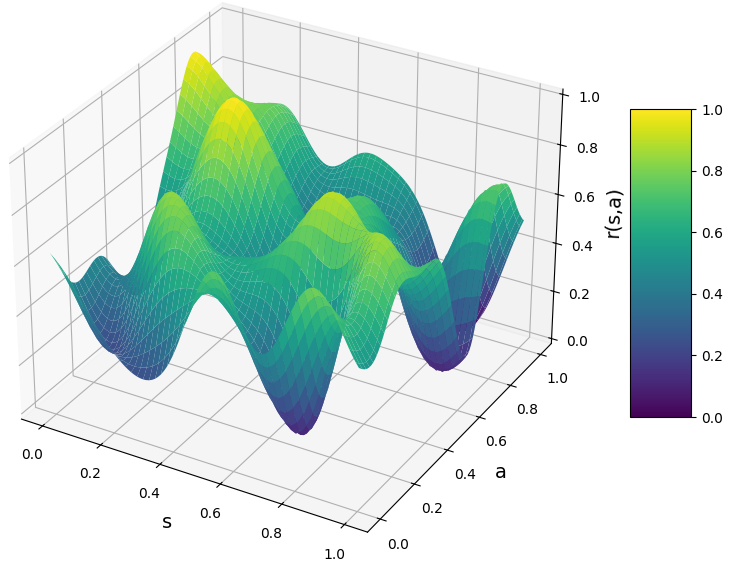}
        \caption{Reward function $r(s,a)$}
        \label{fig:R_RBF}
    \end{subfigure}
    \hfill
    \begin{subfigure}[b]{0.3\textwidth}
        \centering
        \includegraphics[width=\textwidth]{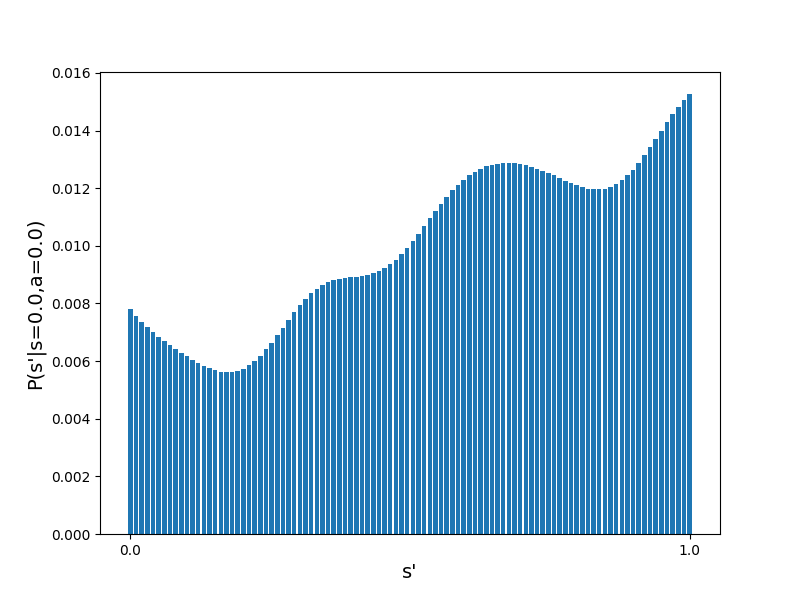}
        \caption{$P(s'|(s=0,a=0)$}
        \label{fig:P_RBF_1}
    \end{subfigure}
    \hfill
    \begin{subfigure}[b]{0.3\textwidth}
        \centering
        \includegraphics[width=\textwidth]{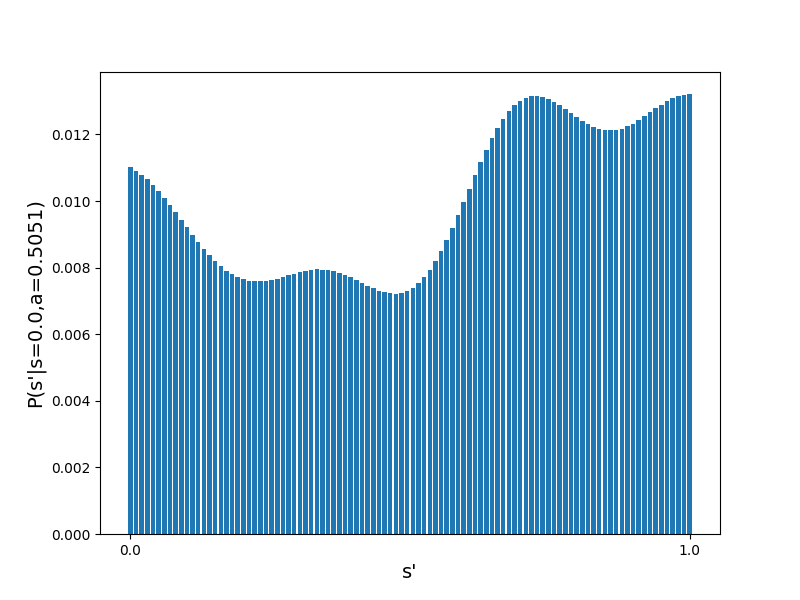}
        \caption{$P(s'|(s=0,a=0.5051)$}
        \label{fig:P_RBF_2}
    \end{subfigure}

    \vspace{\baselineskip} % Add vertical space between rows

    \begin{subfigure}[b]{0.3\textwidth}
        \centering
        \includegraphics[width=\textwidth]{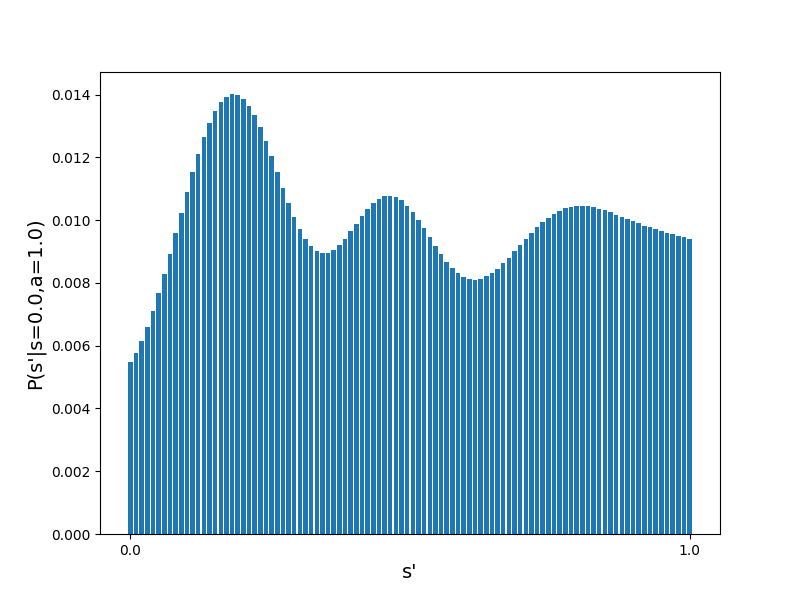}
        \caption{$P(s'| (s=0,a=1)$}
        \label{fig:P_RBF_3}
    \end{subfigure}
    \hfill
    \begin{subfigure}[b]{0.3\textwidth}
        \centering
        \includegraphics[width=\textwidth]{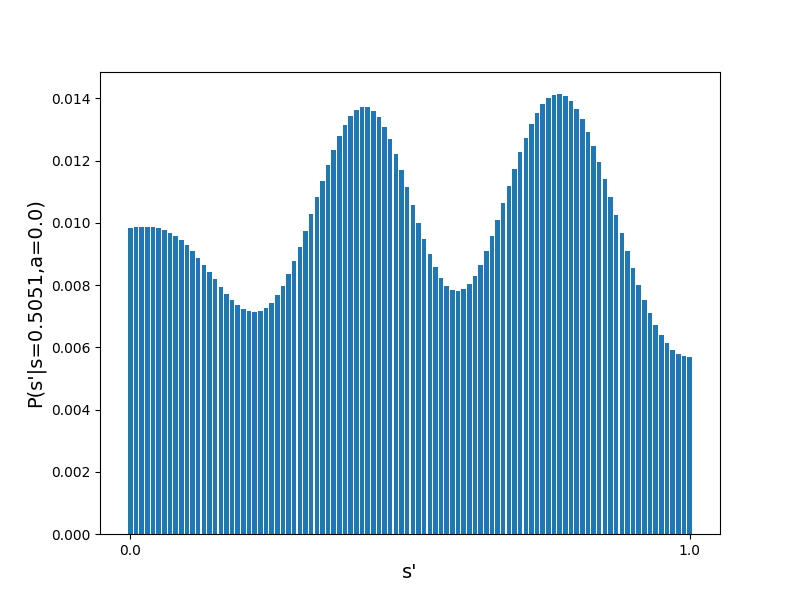}
        \caption{$P(s'| (s=0.5051,a=0)$}
        \label{fig:R_RBF_4}
    \end{subfigure}
    \hfill
    \begin{subfigure}[b]{0.3\textwidth} % Adjust width as needed
        \centering
        \includegraphics[width=\textwidth]{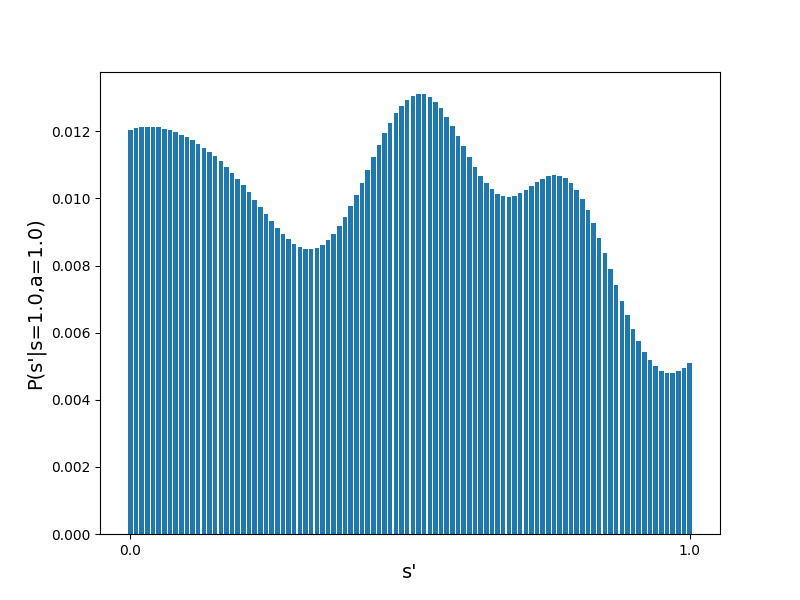}
        \caption{$P(s'|(s=1,a=1)$}
        \label{fig:P_RBF_7}
    \end{subfigure}  
  \caption{Reward and transition probability functions generated by kernel ridge regression using SE Kernel with lengthscale $=0.1$ and $\tau= 0.01$}
  \label{fig:R_P_RBF}
\end{figure}

\begin{figure}[H]
    \centering
    \begin{subfigure}[b]{0.3\textwidth} % Adjust width as needed
        \centering
        \includegraphics[width=\textwidth]{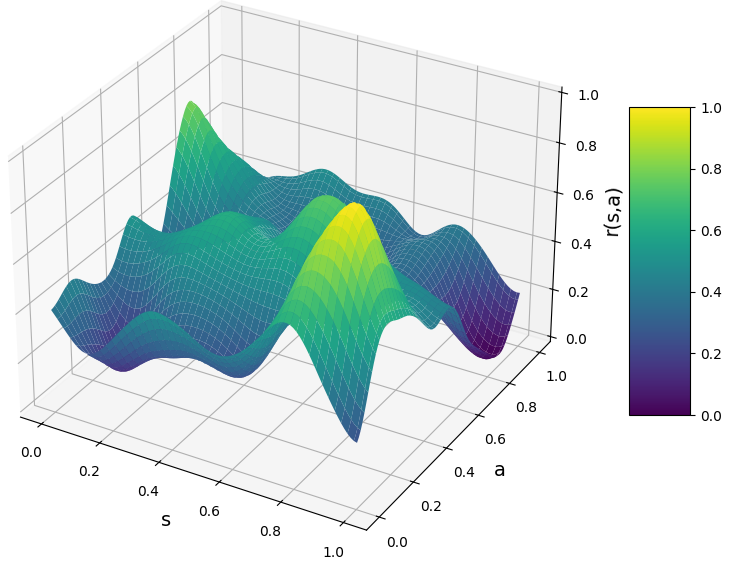}
        \caption{Reward function $r(s,a)$}
        \label{fig:R_Matern2.5}
    \end{subfigure}
    \hfill
    \begin{subfigure}[b]{0.3\textwidth} % Adjust width as needed
        \centering
        \includegraphics[width=\textwidth]{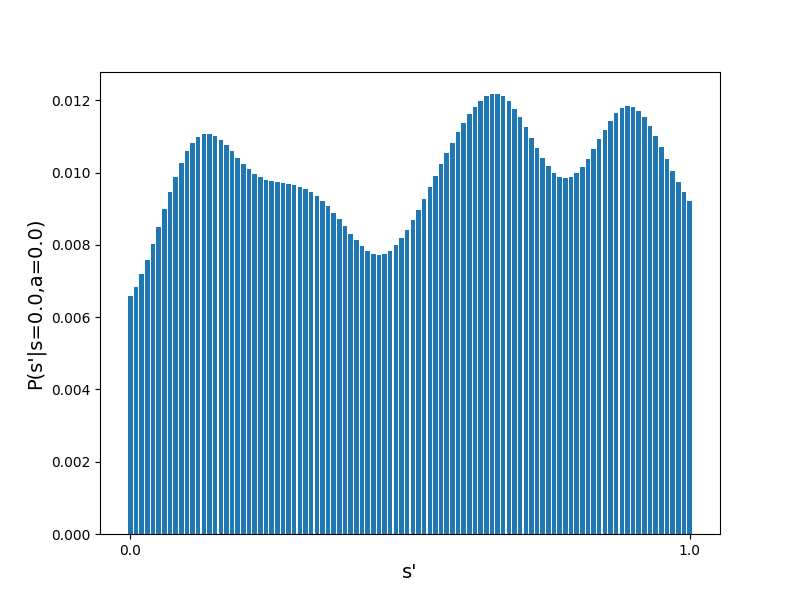}
        \caption{$P(s'|(s=0,a=0)$}
        \label{fig:P_Matern2.5_1}
    \end{subfigure}
    \hfill
    \begin{subfigure}[b]{0.3\textwidth} % Adjust width as needed
        \centering
        \includegraphics[width=\textwidth]{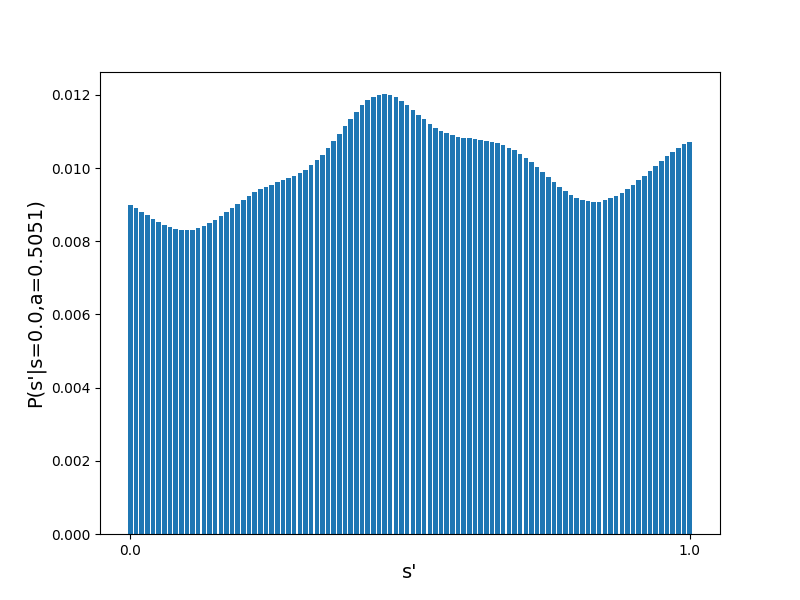}
        \caption{$P(s'|(s=0,a=0.5051)$}
        \label{fig:P_Matern2.5_2}
    \end{subfigure}
\vspace{\baselineskip} % Add vertical space between rows
    \begin{subfigure}[b]{0.3\textwidth} % Adjust width as needed
        \centering
        \includegraphics[width=\textwidth]{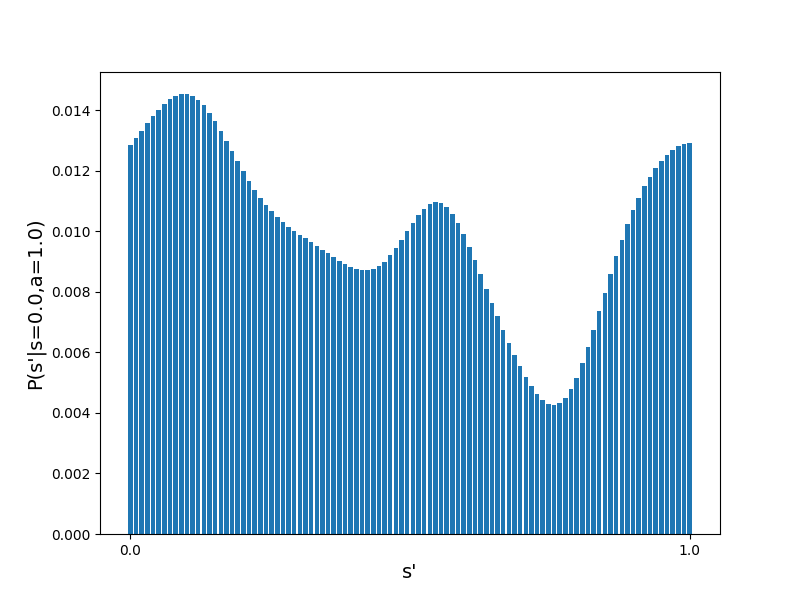}
        \caption{$P(s'|(s=0,a=1)$}
        \label{fig:P_Matern2.5_3}
    \end{subfigure}
    \hfill
    \begin{subfigure}[b]{0.3\textwidth} % Adjust width as needed
        \centering
        \includegraphics[width=\textwidth]{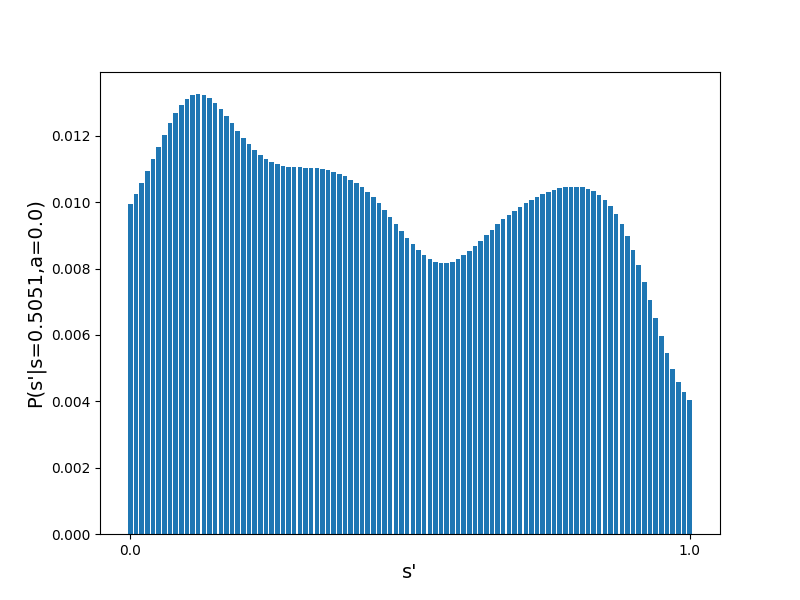}
        \caption{$P(s'|(s=0.5051,a=0)$}
        \label{fig:R_Matern2.5_4}
    \end{subfigure}
    \hfill
    \begin{subfigure}[b]{0.3\textwidth} % Adjust width as needed
        \centering
        \includegraphics[width=\textwidth]{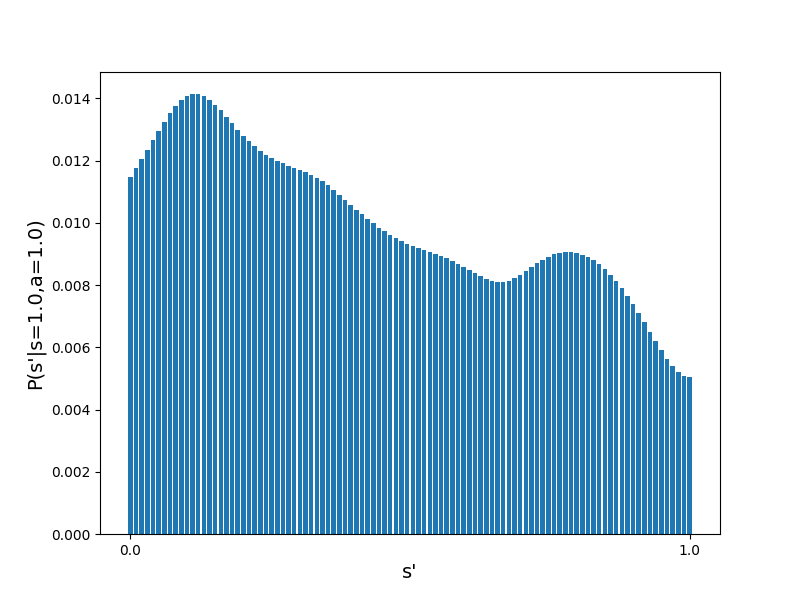}
        \caption{$P(s'|(s=1,a=1)$}
        \label{fig:P_Matern2.5_7}
    \end{subfigure}
    \caption{Reward and transition probability functions generated by kernel ridge regression using Mat{\'e}rn kernel with $\nu=2.5$, lengthscale $=0.1$ and $\tau=0.5$}
    \label{fig:R_P_Matern2.5}
\end{figure}

\begin{figure}[H]
    \centering
    \begin{subfigure}[b]{0.3\textwidth} % Adjust width as needed
        \centering
        \includegraphics[width=\textwidth]{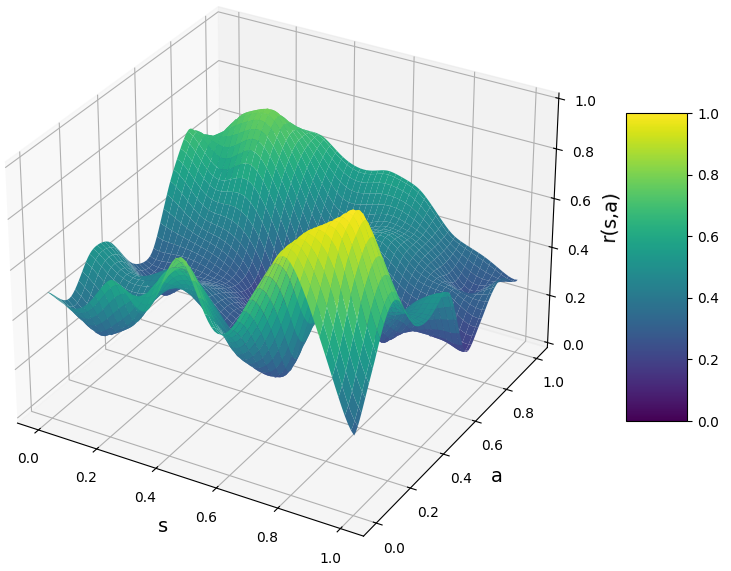}
        \caption{Reward function $r(s,a)$}
        \label{fig:R_Matern1.5}
    \end{subfigure}
    \hfill
    \begin{subfigure}[b]{0.3\textwidth} % Adjust width as needed
        \centering
        \includegraphics[width=\textwidth]{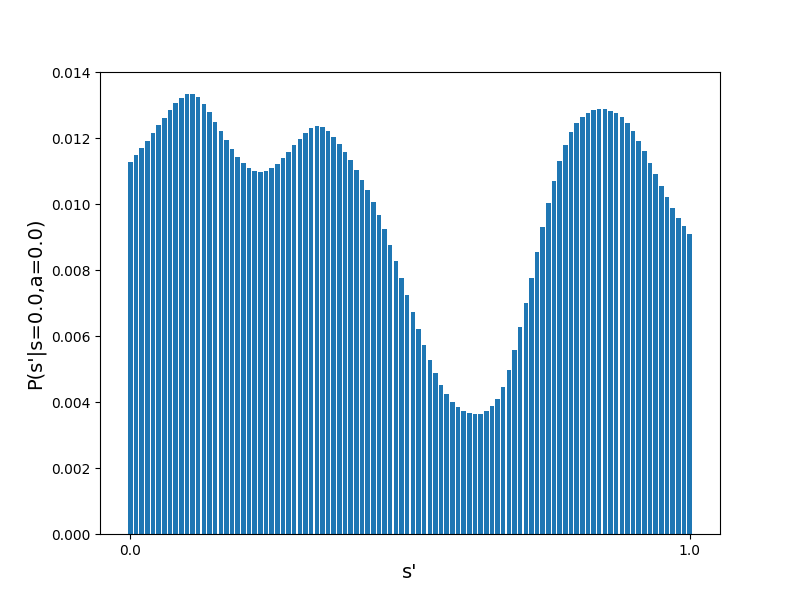}
        \caption{$P(s'|(s=0,a=0)$}
        \label{fig:P_Matern1.5_1}
    \end{subfigure}
    \hfill
    \begin{subfigure}[b]{0.3\textwidth} % Adjust width as needed
        \centering
        \includegraphics[width=\textwidth]{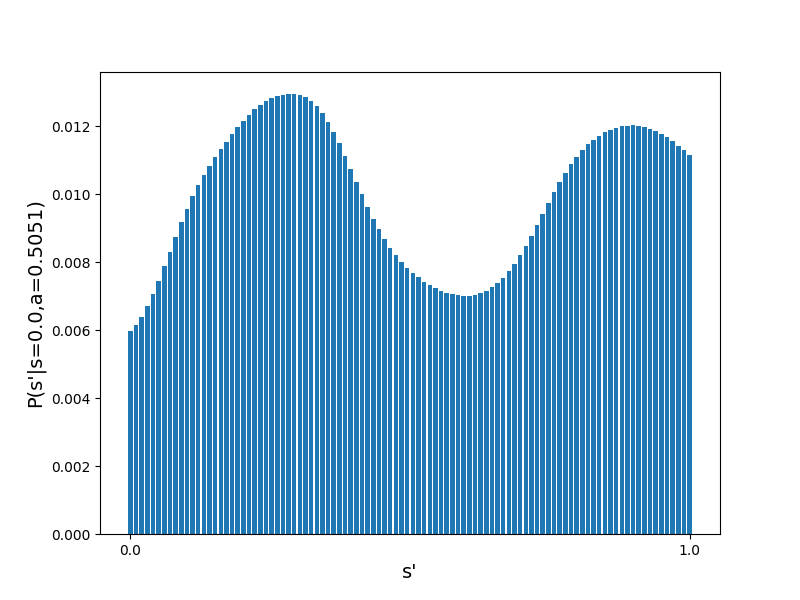}
        \caption{$P(s'|(s=0,a=0.5051)$}
        \label{fig:P_Matern1.5_2}
    \end{subfigure}
\vspace{\baselineskip} % Add vertical space between rows
    \begin{subfigure}[b]{0.3\textwidth} % Adjust width as needed
        \centering
        \includegraphics[width=\textwidth]{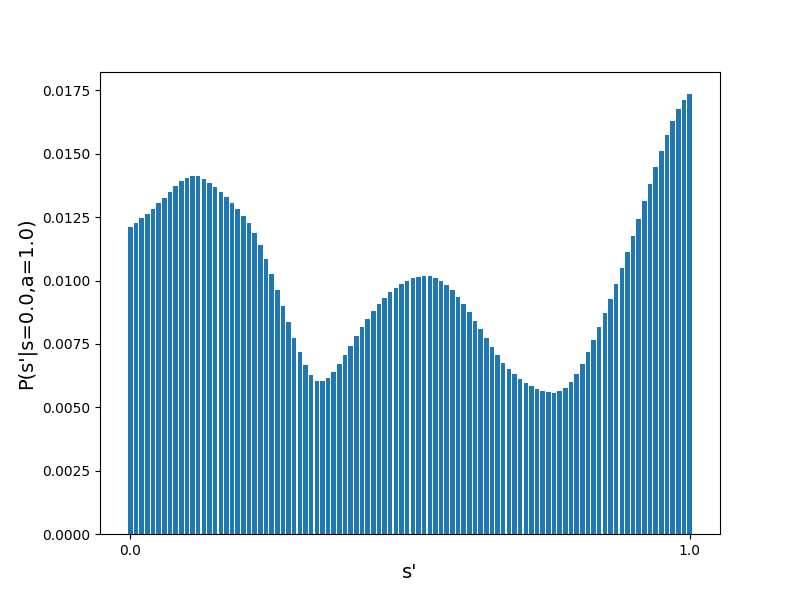}
        \caption{$P(s'|(s=0,a=1)$}
        \label{fig:P_Matern1.5_3}
    \end{subfigure}
    \hfill
    \begin{subfigure}[b]{0.3\textwidth} % Adjust width as needed
        \centering
        \includegraphics[width=\textwidth]{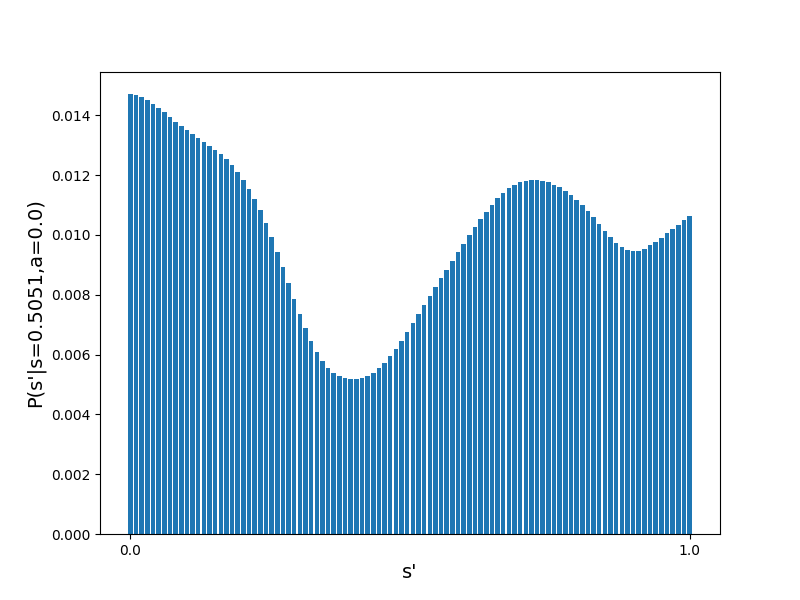}
        \caption{$P(s'|(s=0.5051,a=0)$}
        \label{fig:R_Matern1.5_4}
    \end{subfigure}
    \hfill
    \begin{subfigure}[b]{0.3\textwidth} % Adjust width as needed
        \centering
        \includegraphics[width=\textwidth]{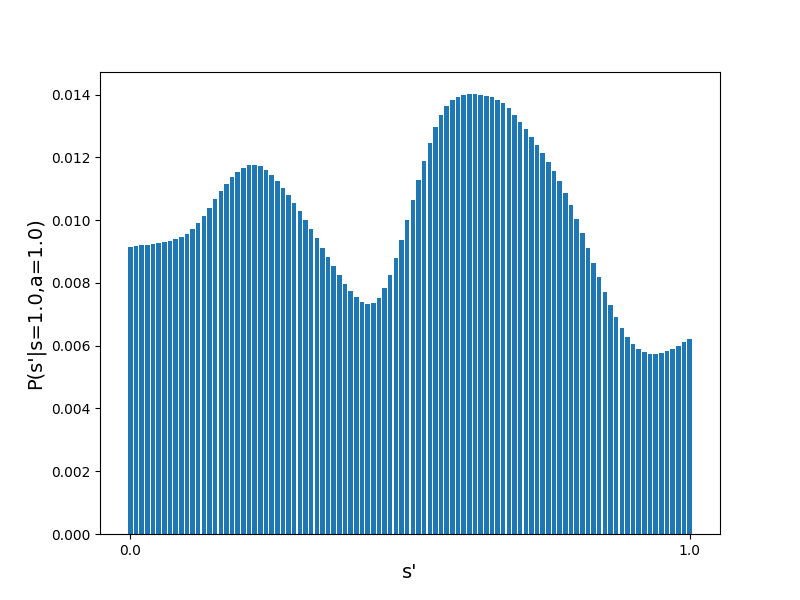}
        \caption{$P(s'|(s=1,a=1)$}
        \label{fig:P_Matern1.5_7}
    \end{subfigure}
    \caption{Reward and transition probability functions generated by kernel ridge regression using Mat{\'e}rn kernel with $\nu=1.5$, lengthscale $=0.1$ and $\tau=0.5$}
    \label{fig:R_P_Matern1.5}
\end{figure}

Our reward function $r$ and transition probability $P$ are arbitrarily chosen functions from an RKHS. For the reward function $r$, we draw a Gaussian Process (GP) sample on a subset of the domain $\Zc$. This subset is generated by sampling a set of evenly spaced points on a $10 \times 10$ grid spanning the range $[0, 1]$ in both dimensions. We then fit kernel ridge regression to these samples and scale the resulting predictions to fit the $[0,1]$ range to obtain $r$. For $P(s'|s,a)$, we similarly draw a GP sample on a subset of the domain $\Zc\times \Sc$, fit kernel ridge regression to these samples, and then shift and rescale for each $z$ to obtain  $P(\cdot|z)$ as a conditional probability distribution. We use the same kernel as the one used in the algorithm. This is a common approach to create functions belonging to an RKHS ~\citep[e.g., see,][]{chowdhury2017kernelized}. Examples of $r$ and $P$ are visualized in Figures \ref{fig:R_P_RBF}, \ref{fig:R_P_Matern2.5} and \ref{fig:R_P_Matern1.5} using Squared Exponential (SE) and Mat{\'e}rn kernels with parameter $\nu=2.5$ and $\nu=1.5$, respectively. For all kernels we use lengthscale of $0.1$. With SE kernel we use $\tau=0.01$, and with Mat{\'e}rn kernels we use $\tau=0.5$. %~\citep[see, e.g., Section 6]{chowdhury2017kernelized}.\\

%\subsection{Matern kernel with smoothness 2.5}

%\subsection{Matern kernel with smoothness 1.5}

\subsection{Implementation and Computational Resources}
%\vspace{-15pt}
For kernel ridge regression, we used Sickit-Learn library \citep{scikit-learn}, which offers robust and efficient tools for implementing and tuning kernel-based machine learning models.
The simulations were executed on a cluster which has $376.2$ GiB of RAM, and an Intel(R) Xeon(R) Gold 5118 CPU running at $2.30$ GHz. The algorithm by \citep{qiu2021reward}, our algorithm without a generative model, and the Greedy Max Variance algorithm typically require approximately 2 minutes of CPU time on average per run. However, our algorithm with a generative model requires around 7 minutes per run due to the cost of increasing the number of exploration episodes by a factor of $H$.
%\vspace{-15pt}
\subsection{Tuning the Confidence Interval Width Multiplier}
\begin{figure}[h]
    \centering
    \begin{subfigure}[b]{0.3\textwidth} % Adjust width as needed
        \centering
        \includegraphics[width=\textwidth]{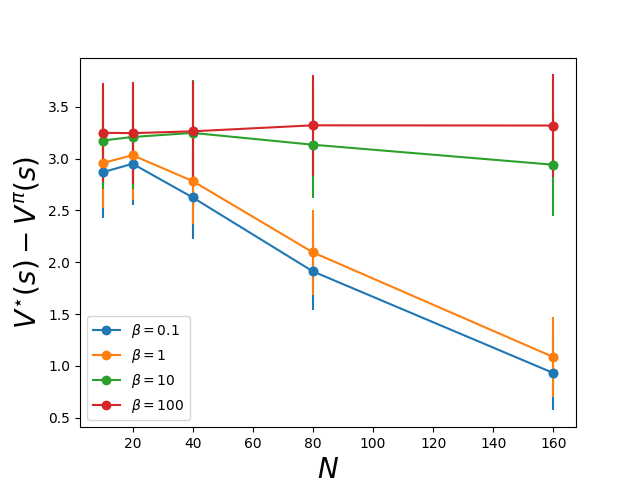}
        \caption{(Qiu et al., 2021)}
        \label{fig:RBF_benchmark_differentbeta}
    \end{subfigure}% <-- No whitespace here
    \begin{subfigure}[b]{0.3\textwidth} % Adjust width as needed
        \centering
        \includegraphics[width=\textwidth]{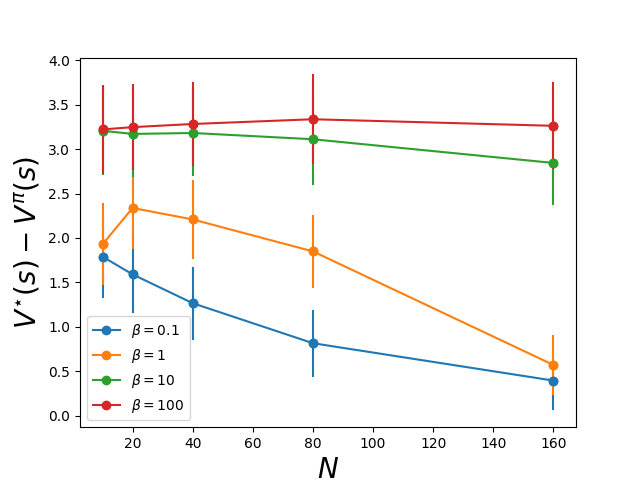}
        \caption{Without generative model}
        \label{fig:/RBF_Maxvariance2_differentbeta}
    \end{subfigure}
     \begin{subfigure}[b]{0.3\textwidth} % Adjust width as needed
        \centering
        \includegraphics[width=\textwidth]{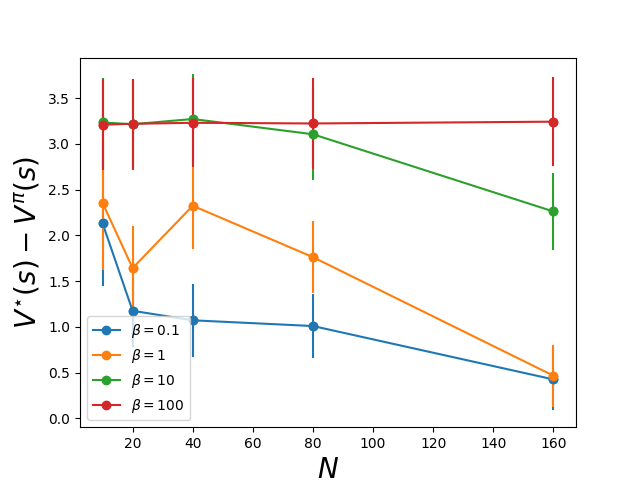}
        \caption{With generative model}
       
    \end{subfigure}% <-- No whitespace here
     \begin{subfigure}[b]{0.3\textwidth} % Adjust width as needed
        \centering
        \includegraphics[width=\textwidth]{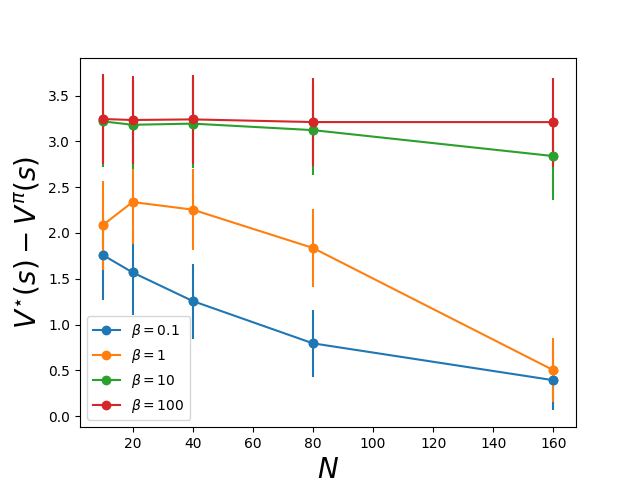}
        \caption{Greedy Max Variance}
        
    \end{subfigure}% <-- No whitespace here
    \caption{Average suboptimality gap plotted against the number of episodes $N$ for different values of $\beta$ in the case of SE kernel.}
    \label{fig:Hyperparams_RBF}
\end{figure}

\begin{figure}[h]
    \centering
    \begin{subfigure}[b]{0.3\textwidth} % Adjust width as needed
        \centering
        \includegraphics[width=\textwidth]{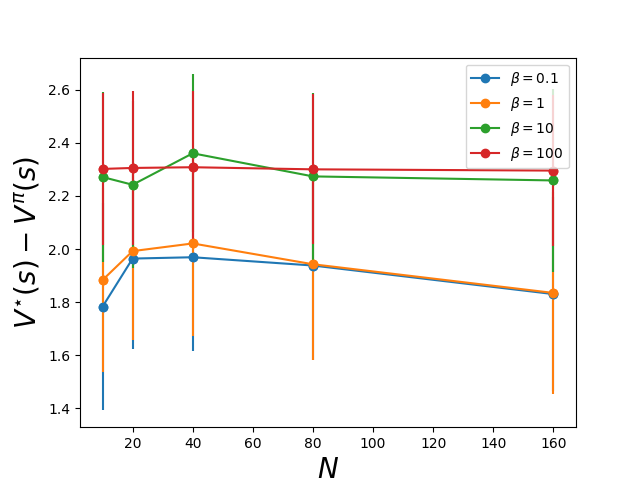}
        \caption{(Qiu et al., 2021)}
        \label{fig:Matern2.5_benchmark_differentbeta}
    \end{subfigure}% <-- No whitespace here
    \begin{subfigure}[b]{0.3\textwidth} % Adjust width as needed
        \centering
        \includegraphics[width=\textwidth]{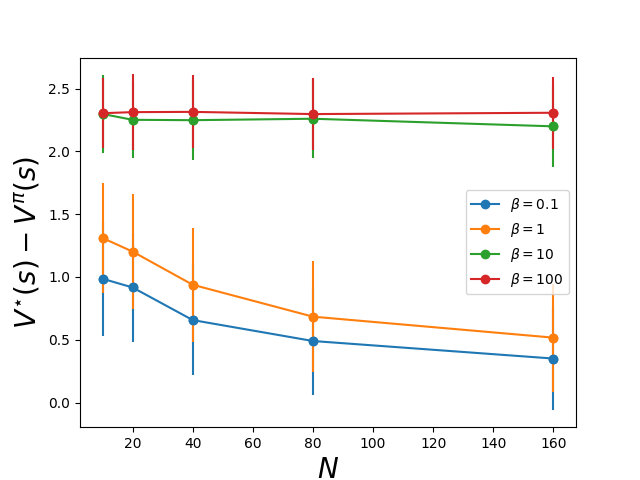}
        \caption{Without generative model}
        \label{fig:/Matern2.5_Maxvariance2_differentbeta}
    \end{subfigure}
         \begin{subfigure}[b]{0.3\textwidth} % Adjust width as needed
        \centering
        \includegraphics[width=\textwidth]{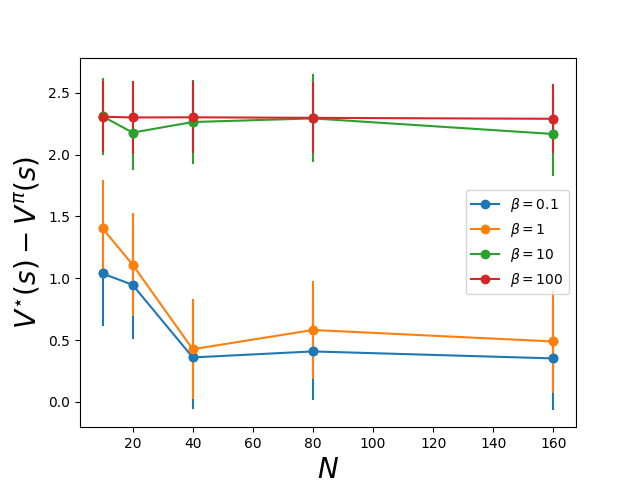}
        \caption{With generative model}
       
    \end{subfigure}% <-- No whitespace here
     \begin{subfigure}[b]{0.3\textwidth} % Adjust width as needed
        \centering
        \includegraphics[width=\textwidth]{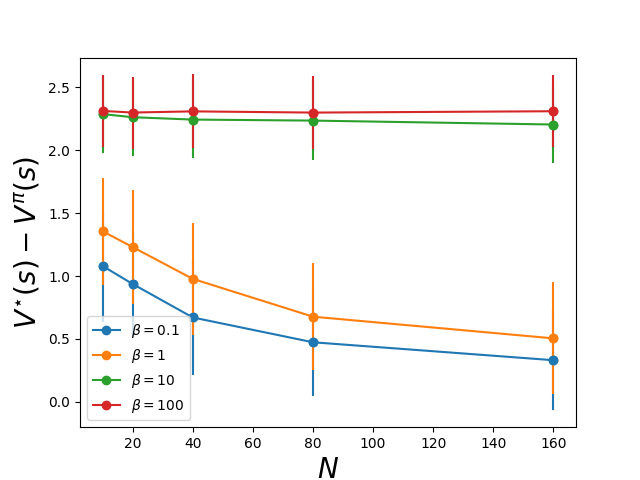}
        \caption{Greedy Max Variance}
        
    \end{subfigure}% <-- No whitespace here
    \caption{Average suboptimality gap plotted against the number of episodes $N$ for different values of $\beta$ in the case of Mat{\'e}rn kernel with $\nu=2.5$.}
    \label{fig:Hyperparams_Matern2.5}
\end{figure}

\begin{figure}[H]
    \centering
    \begin{subfigure}[b]{0.3\textwidth} % Adjust width as needed
        \centering
        \includegraphics[width=\textwidth]{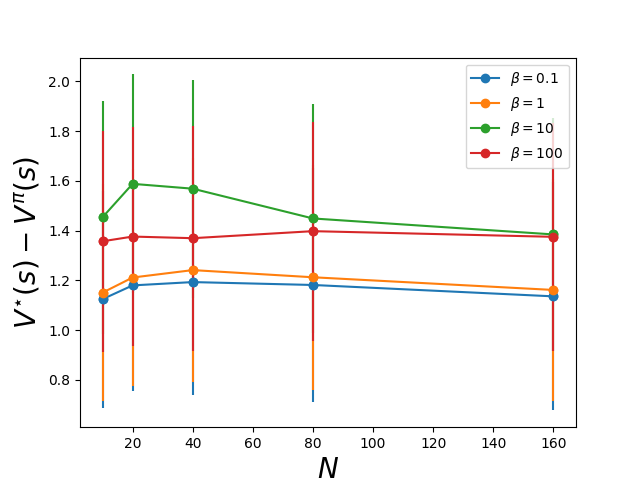}
        \caption{(Qiu et al., 2021)}
        \label{fig:Matern1.5_benchmark_differentbeta}
    \end{subfigure}% <-- No whitespace here
    \begin{subfigure}[b]{0.3\textwidth} % Adjust width as needed
        \centering
        \includegraphics[width=\textwidth]{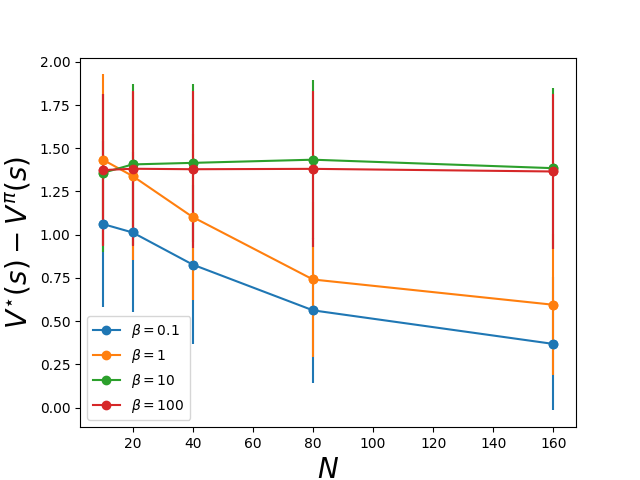}
        \caption{Without generative model}
        \label{fig:/Matern1.5_Maxvariance2_differentbeta}
    \end{subfigure}
         \begin{subfigure}[b]{0.3\textwidth} % Adjust width as needed
        \centering
        \includegraphics[width=\textwidth]{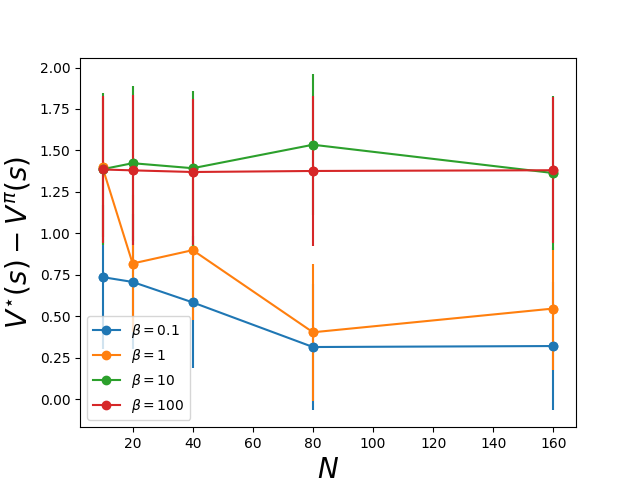}
        \caption{With generative model}
       
    \end{subfigure}% <-- No whitespace here
     \begin{subfigure}[b]{0.3\textwidth} % Adjust width as needed
        \centering
        \includegraphics[width=\textwidth]{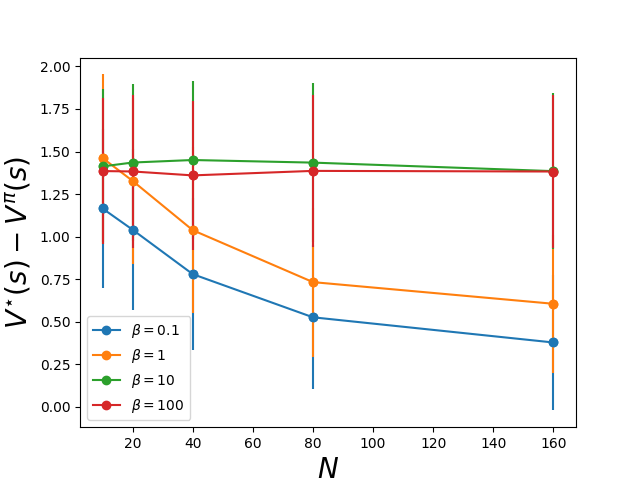}
        \caption{Greedy Max Variance}
        
    \end{subfigure}% <-- No whitespace here
    \caption{Average suboptimality gap plotted against the number of episodes $N$ for different values of $\beta$ in the case of Mat{\'e}rn kernel with $\nu=1.5$.}
    \label{fig:Hyperparams_Matern1.5}
\end{figure}

We perform hyperparameter tuning for the confidence interval width multiplier $\beta$. The theoretical analysis, especially in~\cite{qiu2021reward}, leads to  high values for $\beta$. To ensure a fair comparison between algorithms, we fine-tune  $\beta$ for~\cite{qiu2021reward}, our algorithm without a generative model, our algorithm with generative model and Greedy Max Variance, selecting the best value for each. Figures \ref{fig:Hyperparams_RBF}, \ref{fig:Hyperparams_Matern2.5}, \ref{fig:Hyperparams_Matern1.5}  show the simulation results for various values of $\beta \in [0.1,1,10,100]$ for several kernels. The value $\beta=0.1$ yields the best performance consistently.

\FloatBarrier
\subsection{Repeated Experiments for Different Draws of $r$ and $P$}
\begin{figure}[h]
    \centering
    \begin{subfigure}{0.3\textwidth}
        \centering
        \includegraphics[width=\textwidth]{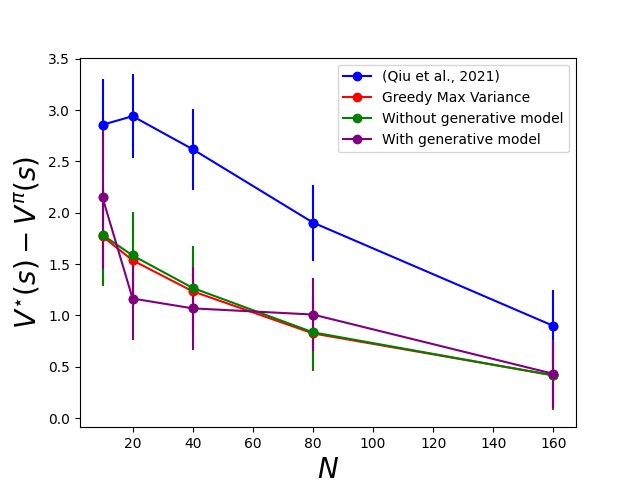}
        \caption{SE Kernel}
        \label{fig:RBF_all_algos_experiment1}
    \end{subfigure}
    %\hspace{0.3em} 
    \begin{subfigure}{0.3\textwidth}
        \centering
        \includegraphics[width=\textwidth]{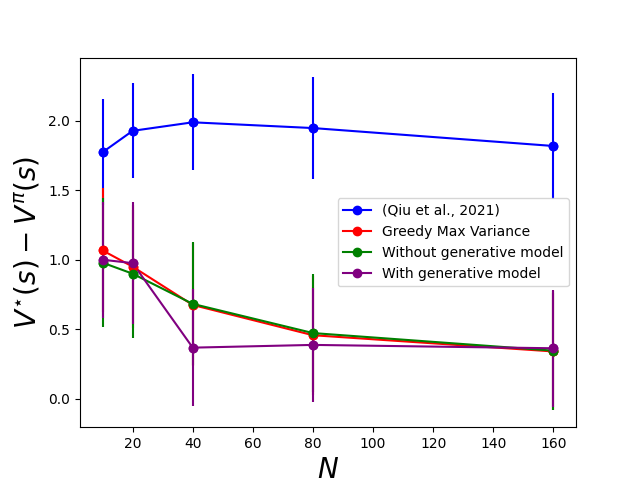} % Replace with the path to your third figure
        \caption{Mat{\'e}rn kernel with $\nu=2.5$}
        \label{fig:Matern2.5_all_algos_experiment1}
    \end{subfigure}
    %\hspace{0.3em} 
    \begin{subfigure}{0.3\textwidth}
        \centering
        \includegraphics[width=\textwidth]{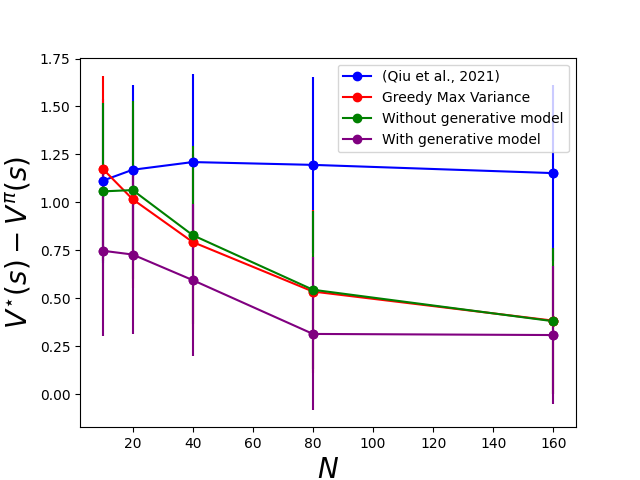} % Replace with the path to your second figure
        \caption{Mat{\'e}rn kernel with $\nu=1.5$}
        \label{fig:Matern1.5_all_algos_experiment1}
    \end{subfigure}
    \caption{Average suboptimality gap plotted against $N$ for repeated experiment 1}
    \label{fig:overallresults_experiment1}
\end{figure}

\begin{figure}[ht]
    \centering
    \begin{subfigure}{0.3\textwidth}
        \centering
        \includegraphics[width=\textwidth]{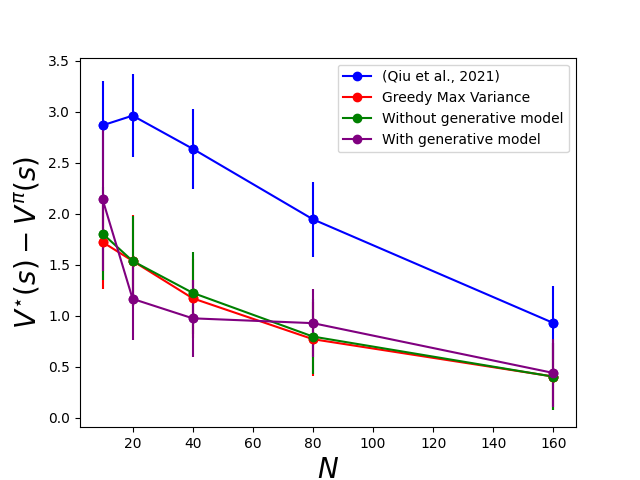}
        \caption{SE Kernel}
        \label{fig:RBF_all_algos_experiment3}
    \end{subfigure}
    %\hspace{0.3em} 
    \begin{subfigure}{0.3\textwidth}
        \centering
        \includegraphics[width=\textwidth]{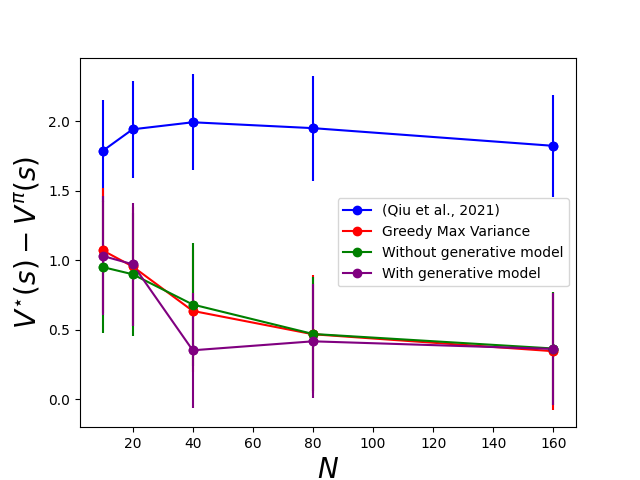} % Replace with the path to your third figure
        \caption{Mat{\'e}rn kernel with $\nu=2.5$}
        \label{fig:Matern2.5_all_algos_experiment3}
    \end{subfigure}
    %\hspace{0.3em} 
    \begin{subfigure}{0.3\textwidth}
        \centering
        \includegraphics[width=\textwidth]{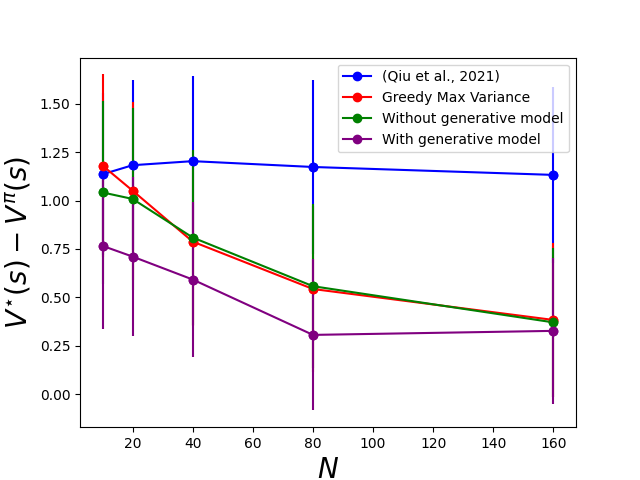} % Replace with the path to your second figure
        \caption{Mat{\'e}rn kernel with $\nu=1.5$}
        \label{fig:Matern1.5_all_algos_experiment3}
    \end{subfigure}
    \caption{Average suboptimality gap plotted against $N$ for repeated experiment 2}
    \label{fig:overallresults_experiment3}
\end{figure}

\begin{figure}[ht]
    \centering
    \begin{subfigure}{0.3\textwidth}
        \centering
        \includegraphics[width=\textwidth]{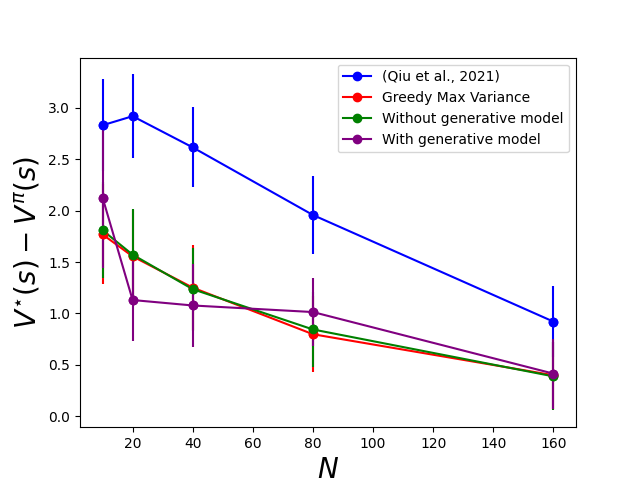}
        \caption{SE Kernel}
        \label{fig:RBF_all_algos_experiment4}
    \end{subfigure}
    %\hspace{0.3em} 
    \begin{subfigure}{0.3\textwidth}
        \centering
        \includegraphics[width=\textwidth]{figures/new_figures/Matern2.5_all_algos_experiment3.png} % Replace with the path to your third figure
        \caption{Mat{\'e}rn kernel with $\nu=2.5$}
        \label{fig:Matern2.5_all_algos_experiment4}
    \end{subfigure}
    %\hspace{0.3em} 
    \begin{subfigure}{0.3\textwidth}
        \centering
        \includegraphics[width=\textwidth]{figures/new_figures/Matern1.5_all_algos_experiment3.png} % Replace with the path to your second figure
        \caption{Mat{\'e}rn kernel with $\nu=1.5$}
        \label{fig:Matern1.5_all_algos_experiment4}
    \end{subfigure}
    \caption{Average suboptimality gap plotted against $N$ for repeated experiment 3}
    \label{fig:overallresults_experiment4}
\end{figure}

To validate the robustness of the results against specific environment realizations, 
we  ran the experiments  three times, with each repetition using different reward and transition probability functions drawn from the RKHS. We kept the hyperparameters (lengthscale, $\tau$, and $\beta$) identical to the ones used in the main paper. The results remained consistent across all repetitions, as shown in Figures \ref{fig:overallresults_experiment1}, \ref{fig:overallresults_experiment3} and \ref{fig:overallresults_experiment4} for different kernels and algorithms.

\vspace{-15pt} % Adjust the value as needed
\newpage
\section{RKHS and Mercer Theorem}\label{appx:rkhs}

Mercer theorem \citep{Mercer1909} provides a representation of the kernel in terms of an infinite dimensional feature map~\citep[e.g., see,][Theorem~$4.49$]{Christmann2008}. Let $\mathcal{Z}$ be a compact metric space and $\mu$ be a finite Borel measure on $\mathcal{Z}$ (we consider Lebesgue measure in a Euclidean space). Let $L^2_\mu(\mathcal{Z})$ be the set of square-integrable functions on $\mathcal{Z}$ with respect to $\mu$. We further say a kernel is square-integrable if
\begin{equation*}
\int_{\mathcal{Z}} \int_{\mathcal{Z}} k^2(z, z') \,d \mu(z) d \mu(z')<\infty.
\end{equation*}

\begin{theorem}
(Mercer Theorem) Let $\mathcal{Z}$ be a compact metric space and $\mu$ be a finite Borel measure on~$\mathcal{Z}$. Let $k$ be a continuous and square-integrable kernel, inducing an integral operator $T_k:L^2_\mu(\mathcal{Z})\rightarrow L^2_\mu(\mathcal{Z})$ defined by
\begin{equation*}
\left(T_k f\right)(\cdot)=\int_{\mathcal{Z}} k(\cdot, z') f(z') \,d \mu(z')\,,
\end{equation*}
where $f\in L^2_\mu(\mathcal{Z})$. Then, there exists a sequence of eigenvalue-eigenfeature pairs $\left\{(\gamma_m, \varphi_m)\right\}_{m=1}^{\infty}$ such that $\gamma_m >0$, and $T_k \varphi_m=\gamma_m \varphi_m$, for $m \geq 1$. Moreover, the kernel function can be represented as
\begin{equation*}
k\left(z, z^{\prime}\right)=\sum_{m=1}^{\infty} \gamma_m \varphi_m(z) \varphi_m\left(z^{\prime}\right),
\end{equation*}
where the convergence of the series holds uniformly on $\mathcal{Z} \times \mathcal{Z}$.
\end{theorem}

According to the Mercer representation theorem~\citep[e.g., see,][Theorem $4.51$]{Christmann2008}, the RKHS induced by~$k$ can consequently be represented in terms of $\{(\gamma_m,\varphi_m)\}_{m=1}^\infty$.

\begin{theorem}(Mercer Representation Theorem) Let $\left\{\left(\gamma_m,\varphi_m\right)\right\}_{i=1}^{\infty}$ be the Mercer eigenvalue-eigenfeature pairs. Then, the RKHS of $k$ is given by
\begin{equation*}
\mathcal{H}_k=\left\{f(\cdot)=\sum_{m=1}^{\infty} w_m \gamma_m^{\frac{1}{2}} \varphi_m(\cdot): w_m \in \mathbb{R},\|f\|_{\mathcal{H}_k}^2:=\sum_{m=1}^{\infty} w_m^2<\infty\right\}.
\end{equation*}
\end{theorem}
Mercer representation theorem indicates that the scaled eigenfeatures $\{\sqrt{\gamma_m}\varphi_m\}_{m=1}^\infty$ form an orthonormal basis for~$\Hc_k$.

\end{document}